\documentclass[pdflatex,sn-vancouver-ay]{sn-jnl}% Vancouver Author Year Reference Style
\usepackage{graphicx}%
\usepackage{multirow}%
\usepackage{amsmath,amssymb,amsfonts}%
\usepackage{amsthm}%
\usepackage{mathrsfs}%
\usepackage[title]{appendix}%
\usepackage{xcolor}%
\usepackage{textcomp}%
\usepackage{manyfoot}%
\usepackage{booktabs}% 
\usepackage{url}%
\usepackage{xpinyin}
\usepackage{longtable} 
\usepackage{algorithm}%
\usepackage{algorithmicx}%
\usepackage{algpseudocode}
\usepackage{listings}%
\usepackage{array}
\usepackage{multirow}
\usepackage{hyperref}
\usepackage{geometry}
\usepackage{rotating}
\usepackage{adjustbox}
\usepackage{siunitx}
\usepackage{lscape} 
\usepackage[table,xcdraw]{xcolor} % Include xcolor package for table coloring
\definecolor{lightblue}{rgb}{0.678, 0.847, 0.902}
\definecolor{lightred}{rgb}{1.0, 0.678, 0.678}
\usepackage{fontspec}  % For font selection
\usepackage{xeCJK}     % For CJK support in XeTeX

\theoremstyle{thmstyleone}%
\theoremstyle{thmstyletwo}%

\theoremstyle{thmstylethree}%
\newtheorem{definition}{Definition}%

\begin{document}

% \title[Building from Scratch: Human-AI Integration for Multilingual Legal Terminology Mapping]{Building from Scratch: Human-AI Integration for Multilingual Legal Terminology Mapping}
\title[Building from Scratch: A Multi-Agent Framework with Human-in-the-Loop for Multilingual Legal Terminology Mapping]{Building from Scratch: A Multi-Agent Framework with Human-in-the-Loop for Multilingual Legal Terminology Mapping}

\author[1]{\fnm{Lingyi} \sur{Meng}}

\author[2]{\fnm{Maolin} \sur{Liu}}
\author*[2]{\fnm{Hao} \sur{Wang}}\email{wang-hao@shu.edu.cn}
\author[1]{\fnm{Yilan} \sur{Cheng}}

\author[2]{\fnm{Qi} \sur{Yang}}

\author[2]{\fnm{Idlkaid} \sur{Mohanmmed}}

\affil[1]{\orgdiv{School of Foreign Languages}, \orgname{East China Normal University}, \orgaddress{\state{Shanghai}, \country{China}}}

\affil[2]{\orgdiv{School of Computer Engineering and Science}, \orgname{Shanghai University}, \orgaddress{\state{Shanghai}, \country{China}}}

\abstract{Accurately mapping legal terminology across languages remains a significant challenge, especially for language pairs like Chinese and Japanese, which share a large number of homographs with different meanings. Existing resources and standardized tools for these languages are limited. To address this, we propose a human-AI collaborative approach for building a multilingual legal terminology database, based on a multi-agent framework. This approach integrates advanced large language models (LLMs) and legal domain experts throughout the entire process—from raw document preprocessing, article-level alignment, to terminology extraction, mapping, and quality assurance. Unlike a single automated pipeline, our approach places greater emphasis on how human experts participate in this multi-agent system. Humans and AI agents take on different roles: AI agents handle specific, repetitive tasks, such as OCR, text segmentation, semantic alignment, and initial terminology extraction, while human experts provide crucial oversight, review, and supervise the outputs with contextual knowledge and legal judgment. We tested the effectiveness of this framework using a trilingual parallel corpus comprising 35 key Chinese statutes, along with their English and Japanese translations. The experimental results show that this human-in-the-loop, multi-agent workflow not only improves the precision and consistency of multilingual legal terminology mapping but also offers greater scalability compared to traditional manual methods. Additionally, we observed that several open-source large language models performed exceptionally well in legal terminology extraction, demonstrating their cost-effectiveness and potential for sustainable applications in multilingual legal natural language processing (NLP). Finally, the open, extensible platform we developed supports continuous expert curation and can be easily integrated into various legal translation, research, and AI-powered knowledge management tools.\footnote{The data resource can be found in \url{https://www.chineselawtranslation.com}.}\footnote{This paper has been accepted in Artificial Intelligence and Law.}}
%准确映射不同语言间的法律术语仍然是一项重大挑战，对于中文和日语这类存在大量同形异义字的语言对而言尤为如此。目前针对这些语言的现有资源和标准化工具十分有限。为解决此问题，我们提出了一种人机协同构建多语种法律术语数据库的方法，该方法基于多智能体框架。此方法将先进的大语言模型和法律领域专家整合到整个流程中——从原始文档预处理、条文级对齐，到术语提取、映射和质量保证。与单一的自动化流程不同，我们的方法更侧重于人类专家在此多智能体系统中的参与方式。人类与AI智能体承担不同角色：AI智能体处理具体、重复性的任务，如OCR、文本分割、语义对齐和初始术语提取；而人类专家则凭借其背景知识和法律判断，提供关键的监督、审阅和指导作用。我们使用一个包含35部中国关键法律及其英、日文译本的三语平行语料库测试了该框架的有效性。实验结果表明，这种"人在回路"的多智能体工作流不仅提高了多语种法律术语映射的精确度和一致性，而且相比传统人工方法具有更强的可扩展性。此外，我们观察到多个开源大语言模型在法律术语提取任务中表现优异，展现了其在多语种法律自然语言处理应用中成本效益和可持续应用的潜力。最后，我们开发的开放、可扩展平台支持专家的持续整理工作，并能轻松集成到各种法律翻译、研究以及AI驱动的知识管理工具中。




\maketitle

\section{Introduction}\label{sec1}
As international legal frameworks, trade agreements, and cross-border business activities continue to expand, the importance of clear and reliable legal communication has grown correspondingly. Based on our practical experience and the accounts of many legal professionals, significant obstacles remain when navigating the diverse terminologies and legal concepts that characterize different jurisdictions \citep{vsarcevic2000legal,terral2004empreinte}. Many established translation approaches often struggle to convey the full depth and context-specific meanings of legal terms, particularly when subtle distinctions or system-specific usages are involved \citep{cao2007translating,NAVEEN2024110878}. These gaps are not merely theoretical; in practice, they frequently lead to misinterpretations, compliance risks, and inefficiencies in cross-border legal matters \citep{Prieto-Ramos04032021,Qu2015,Zhao2023NewAdvances}. As a result, robust and accurate mapping of legal terminology across jurisdictions is essential—not only as a scholarly endeavor, but as a practical necessity, especially for languages such as Chinese and Japanese, where unique historical and cultural legacies add further complexity to achieving conceptual equivalence and mutual understanding \citep{Kozanecka2018141162}.

%随着国际法律框架、贸易协定和跨境商业活动的持续扩展，清晰可靠的法律沟通的重要性也相应增长。根据我们的实践经验以及众多法律专业人士的陈述，在应对不同司法管辖区所特有的多样化术语和法律概念时，仍然存在重大障碍 \citep{vsarcevic2000legal,terral2004empreinte}。许多既定的翻译方法常常难以传达法律术语的全部深度和特定语境下的含义，尤其是在涉及微妙区别或体系特定用法时 \citep{cao2007translating,NAVEEN2024110878}。这些差距不仅仅是理论上的；在实践中，它们常常导致跨境法律事务中的误解、合规风险以及效率低下 \citep{Prieto-Ramos04032021,Qu2015,Zhao2023NewAdvances}。因此，构建跨司法管辖区的稳健且准确的法律术语映射至关重要——这不仅是一项学术努力，更是一种实际需要，特别是对于中文和日语等语言，其独特的历史和文化传统为实现概念对等和相互理解增添了进一步的复杂性 \citep{Kozanecka2018141162}。

Developing a Multilingual Legal Terminology Database (MLTD) is an inherently complex and labor-intensive process, often spanning years of sustained effort to keep pace with ever-evolving legal frameworks. As noted by \cite{Chiocchetti2023}, the creation of such a platform requires meticulous needs analysis, systematic term extraction, and rigorous quality assurance—each step involving specialized workflows and distinct professional roles. This endeavor necessitates close collaboration among language mediators (translators and interpreters), legal scholars, and software engineers. 
%开发多语种法律术语数据库（MLTD）是一项复杂且 inherently 劳动密集型的过程，通常需要持续数年的努力才能跟上不断演进的法律框架。正如 \cite{Chiocchetti2023} 所指出的，创建这样一个平台需要 meticulous 的需求分析、系统性的术语提取和 rigorous 的质量保证——每一步都涉及专业化的工作流程和不同的职业角色。这项事业通常需要语言中介者（笔译员、口译员）、法律学者和软件工程师的紧密协作。

Recent advances in large language models (LLMs) have begun to address some of these longstanding challenges. AI-driven legal research tools now enable faster, more accurate, and efficient analysis of vast legal datasets across multiple jurisdictions, providing a powerful complement to traditional human expertise. Applications of LLMs in legal informatics already span automated term extraction \citep{breton2025llm_lex_extraction}, judgment summarization \citep{Gao2025}, and legal question answering \citep{hu-etal-2025-fine}. Furthermore, the integration of multi-agent systems \citep{yao2023react,10.1145/3672459}, in which multiple AI agents coordinate their efforts using advanced LLMs, is showing promise in handling complex tasks such as domain-specific document translation \citep{wu2025perhapshumantranslationharnessing} and collaborative legal consultation \citep{cui2024chatlawmultiagentcollaborativelegal}.

This study addresses the fundamental challenges of multilingual legal terminology extraction, alignment, and management—not by generating isolated monolingual term lists, but by enabling context-sensitive, cross-lingual mapping among Chinese, Japanese, and English legal systems. Leveraging a large-scale, article-aligned corpus of major Chinese legal codes and their official translations, we introduce a comprehensive workflow that tightly integrates advanced language models with validation by domain experts. 

Our method moves beyond traditional automated extraction: it supports scalable and high-quality alignment of legal terms, combined with rigorous quality assurance mechanisms that function reliably across multiple jurisdictions. By combining computational linguistics techniques with specialized legal knowledge, this research establishes a practical, extensible foundation for sustainable terminology management. The resulting framework facilitates not only legal translation and comparative law research, but also the development of intelligent, multilingual legal information systems with global reach.

In summary, the main contributions of this paper are as follows:
\begin{enumerate}
\item \textbf{Construction of a large-scale multilingual legal parallel corpus}: We have constructed a large-scale, article-aligned parallel corpus comprising 35 foundational Chinese legal codes alongside their high-quality official English and Japanese translations. Through a multi-agent framework, we achieve article-level alignment and ensure both cross-lingual consistency and professional translation standards across diverse legal domains.
% \textbf{大规模多语种法律平行语料库构建}：我们构建并对齐了35部中国基础法律及其高质量英、日译本，依托多智能体框架实现条款级精准对齐，保障跨语一致性和多法域的专业翻译标准。

\item \textbf{Innovative methodology for multilingual terminology mapping}: We present a multi-stage mapping pipeline that combines multi-agent collaborative term extraction with expert validation. This approach enables precise identification, alignment, and contextual mapping of legal terms across Chinese, English, and Japanese, maintaining high semantic fidelity and domain professionalism.
%\textbf{创新的多语法律术语抽取方法}：提出系统化的术语抽取流程，将多智能体协同与专家验证相结合，实现中、英、日三语法律术语的精确识别、对齐与语境映射，兼顾语义一致性与领域规范性。

\item  \textbf{Unified human-AI reference framework for quality assessment}:
We establish a comprehensive five-dimensional evaluation scheme, spanning Coverage, Consistency, Completeness, Professionalism, and Translation Quality, supported by 17 specialized sub-criteria with the assistance of legal linguists and computer scientists. This unified framework is applicable to both human and machine assessment, ensuring consistent, robust, and professional quality assurance for multilingual legal terminology resources.

%\textbf{系统性人机协同术语质量评估框架}：构建覆盖覆盖度、一致性、完整性、专业性、翻译质量五大维度、共17项细分指标的综合评测体系。该框架融合算法评测、大模型自动化与专家人工复核，实现多语法律术语库的量化与专业化质量保障。

\item \textbf{Development of an open, AI-compatible termbase:} 
We have developed a next-generation terminology platform that supports both human users and AI (LLM) models in accessing, retrieving, and utilizing multilingual legal terminologies. Unlike traditional termbases, our platform offers open access, dynamic updates, and interfaces optimized for seamless integration with large language models as well as human experts. This enables more efficient, accurate, and scalable use of legal terminology in research, legal translation, and AI-driven applications, which will be accessible online in the near future.
%\textbf{开放式动态术语管理平台开发}：开发了支持术语实时检索、协作共建、跨法域追溯和质量监控的在线平台，将评测体系集成于平台，实现高质量多语法律术语资源的开放获取、持续完善和实际应用部署。

\end{enumerate} 
\section{Problem Definitions}

The central objective of this study is to develop a Multilingual Legal Terminology Database (MLTDB) that systematically maps legal terminology across Chinese, English and Japanese. Our data sources include a wide range of web-based legal corpora, prioritizing authoritative Chinese legal texts. In keeping with recent scholarship~\citep{Melby2012, Steurs2015, Bowker2015, Chiocchetti2023}, we begin by clarifying several core concepts and definitions:
%本研究的核心目标是设计并构建一个多语种法律术语数据库（MLTDB），系统性地涵盖中文、英文和日文三种法律术语。数据来源包括丰富的互联网法律语料，重点聚焦于权威的中文法律文本。结合近期相关学术研究~\citep{Melby2012, Steurs2015, Bowker2015, Chiocchetti2023}，我们首先对若干核心概念和术语作如下界定：

\begin{definition}[\textbf{Multilingual Legal Terminology Database, MLTDB}]
A Multilingual Legal Terminology Database (MLTDB) is a terminology database designed for the legal profession, containing legal terminology in two or more languages. Its core function is to facilitate the reliable mapping, alignment, and consistency management of legal terms across linguistic and jurisdictional boundaries~\citep{Chiocchetti2023}.
\end{definition}
%MLTDB是一种术语数据库（TDB），其内容覆盖两种及以上语言的法律术语，面向法律行业的实际需求。其核心功能是实现法律术语在不同语言与法域之间的可靠映射、对齐与一致性管理~\citep{Chiocchetti2023}。

\begin{definition}[\textbf{Terminology \& Term Entry}]
A \emph{term} is a linguistic designation representing a general concept within a specialized domain.\footnote{ISO 1087:2019, Clause 3.4.2.} For example, ``laser printer'', ``planet'', and ``pacemaker'' are all terms within their respective domains. In a termbase, a \emph{term entry} refers to a single record that gathers all terminological data related to one specific concept, as defined in ISO 26162:2023.
\end{definition}
%``术语''是指在特定领域内表达一般概念的语言标记。\footnote{ISO 1087:2019，第3.4.2条} 例如，``激光打印机''``行星''``起搏器''均为各自领域的术语。在术语库中，``术语条目''指的是关于某一特定概念所汇集的全部术语数据（ISO 26162:2023）。

\begin{definition}[\textbf{TermBank}]
A \emph{TermBank} is a large-scale, thematically organized, and strictly managed repository of standardized terminology, typically maintained by authoritative institutions. It serves as a centralized resource to ensure terminological consistency, interoperability, and professional accessibility in translation and legal practice~\citep{Bowker2015}. Specifically, a TermBank focuses on upholding industry standards and providing reliable term support for multilingual and cross-cultural legal applications.
\end{definition}
%术语库是一个大规模、主题化且严格管理的标准化术语资源库。通常由权威机构维护，术语库作为一个集中的资源，确保在翻译和法律实践中术语使用的一致性、互操作性和专业可访问性~\citep{Bowker2015}。术语库的重点在于维护行业标准，为多语种和跨文化应用提供一致的术语支持，特别是在法律领域中的应用。

\begin{definition}[\textbf{TermBase}]
A \emph{TermBase} is a computer-based database that stores structured information about domain-specific concepts and their corresponding designations. Entries are systematically enriched with metadata and organized according to concept-driven taxonomies, supporting advanced linguistic and legal analysis~\citep{Melby2012, Steurs2015}. A TermBase emphasizes the detailed documentation of terms, providing rich background information to support specialized translation and legal practice within a specific domain.
\end{definition}

\begin{definition}[\textbf{Users of TermBase}]
Users of a TermBase encompass a wide range of stakeholders: translators, interpreters, legal professionals, legislative drafters, public administrators, international organization staff, and the general public. Increasingly, IT specialists in NLP, machine learning, Semantic Web, and AI also rely on terminological resources to develop and enhance intelligent tools for legal and technical domains~\citep{Chiocchetti2023}.
\end{definition}
%术语基础库的用户涵盖广泛：包括译者、口译员、法律从业者、立法起草人、公共管理人员、国际组织成员及普通公众。随着技术发展，NLP、机器学习、语义网和AI等领域的IT专家，也越来越多地依赖术语资源来开发和优化法律及技术领域的智能工具~\citep{Chiocchetti2023}。

\begin{definition}[\textbf{Quality Management}]
Quality management refers to the set of policies, objectives, and procedures that ensure the reliability and effectiveness of a terminology database. In practice, MLTDBs are embedded within multilingual legal communication workflows and must comply with the overarching quality standards of their host organizations~\citep{drewer2017terminologiemanagement}.
\end{definition}
%质量管理指确保术语数据库可靠性和效能的一系列政策、目标与流程。实际中，MLTDB 通常嵌入多语法律交流流程，须遵循所在机构的整体质量标准~\citep{drewer2017terminologiemanagement}。

This research aims to develop a termbase optimized for broad deployment in AI-driven legal research and applications. To this end, the preferred architecture is concept-oriented, highly indexable, and context-sensitive. Recognizing the expanding role of intelligent agents and large language models, we explicitly extend the user base beyond human practitioners to include autonomous systems. The termbase is thus designed to support both human and AI users, enabling context-aware retrieval and machine-friendly interfaces that maximize the utility of its legal knowledge resources~\citep{speranza-etal-2020-linguistic}.
%本研究致力于开发一个面向AI驱动法律研究和应用、架构以概念为中心、高可检索性、语境敏感的术语基础库。考虑到智能体和大型语言模型作用的日益扩大，术语库的用户对象不仅包括人工实践者，还专门拓展到自主智能系统。所构建术语库既支持人工用户，也面向AI用户，能实现语境感知检索和高效机器接口，最大化其法律知识资源的应用价值~\citep{speranza-etal-2020-linguistic}。

\section{Related Works}
\subsection{Termbase Construction} 
Terminology databases (termbases) function as digital infrastructures for the structured storage and retrieval of specialized terms and their associated metadata, providing essential support for cross-linguistic knowledge management in law and other fields~\citep{schmitz2017terminologiemanagement}. Typically concept-oriented in design, termbases integrate terms, definitions, language information, domain classifications, and contextual data to promote consistency in translation, knowledge standardization, and efficient multilingual content creation.
%术语数据库（termbase）作为结构化存储与检索专业术语及其相关元数据的数字基础设施，为跨语言知识管理，特别是在法律等领域，提供了关键支撑~\citep{schmitz2017terminologiemanagement}。术语库通常采用以概念为核心的设计理念，将术语、定义、语言信息、领域分类和语境等内容有机整合，以促进翻译一致性、知识标准化和多语内容的高效生产。

The development of a Multilingual Legal Terminology Database (MLTDB) synthesizes the traditions of classical lexicography—emphasizing systematic, dictionary-based language description~\citep{atkins2008oxford}—with modern termbase engineering, which is inherently concept-driven, domain-specific, and multilingual~\citep{Chiocchetti2023}. Lexicographic rigor ensures linguistic precision and richness, while the engineering approach enables the interoperability and functional detail needed for legal applications across jurisdictions.
%多语种法律术语数据库（MLTDB）的开发融合了传统词典学——注重系统性、基于词典的语言描述~\citep{atkins2008oxford}——和现代术语库工程的理念，后者本身即以概念为驱动，具有领域专属性与多语特征~\citep{Chiocchetti2023}。词典学的严谨性保障了术语的语言丰富性与精准性，而工程化视角则为跨法域法律应用提供了所需的互操作性和功能细粒度。

International organizations and government agencies have set authoritative standards for legal, scientific, and technical terminology management by developing large-scale termbases with rigorous editorial workflows. Notable examples include the United Nations' UNTERM\footnote{\url{https://unterm.un.org/unterm2/}}, the European Union's IATE\footnote{\url{https://iate.europa.eu/}}, and Canada's Termium Plus\textsuperscript{\textregistered}\footnote{\url{https://www.btb.termiumplus.gc.ca/}}.
%国际组织和政府机构通过建立大型术语库、实施严格编辑流程，为法律、科学及技术领域的术语管理树立了权威标准。代表性案例包括联合国的 UNTERM\footnote{\url{https://unterm.un.org/unterm2/}}、欧盟的 IATE\footnote{\url{https://iate.europa.eu/}} 以及加拿大的 Termium Plus\textsuperscript{\textregistered}\footnote{\url{https://www.btb.termiumplus.gc.ca/}}。

\begin{table}[h]
\centering
\caption{Types of Institutions and Their Representative Terminology Databases}
\begin{tabular}{>{\raggedright\arraybackslash}m{2cm}>{\raggedright\arraybackslash}m{4cm}>{\raggedright\arraybackslash}m{4cm}}
\toprule
\textbf{Institution Type} & \textbf{Representative Database} & \textbf{Characteristics} \\
\midrule
International Organization & United Nations UNTERM, EU IATE & Multilingual, large-scale (IATE covers 26 languages with 935,000 entries) \\
\midrule
Government Agency & Canada Termium Plus\textsuperscript{\textregistered} & Standardized legal and administrative terminology \\
\midrule
Research Institution & EURAC Research (bistro)& Interdisciplinary terminology collaboration \\
\midrule
Enterprise & Microsoft Language Portal & Technical terminology integrated into product ecosystem \\
\bottomrule
\end{tabular}
\end{table}

Theoretical foundations for terminology management have been well established by scholars such as~\cite{sager1990terminology, cabre1998terminology, schmitz2012standards, schmitz2017terminologiemanagement}, who clarified distinctions between termbanks and termbases and advanced methodologies for term extraction. Modern techniques have evolved from early corpus-based statistical analyses (such as co-occurrence frequency) to more sophisticated automated approaches—including rule-based engines and deep learning—that have greatly improved the extraction of multi-word terms (MWTs).
%术语管理的理论基础已由 \cite{sager1990terminology, cabre1998terminology, schmitz2012standards, schmitz2017terminologiemanagement} 等学者系统奠定，对术语库与术语基础库的区分进行了明确界定，并推动了术语抽取方法的不断演进。现代技术已从早期基于语料库的统计分析（如共现频率）发展到更加自动化的解决方案，包括基于规则的引擎和深度学习方法，大幅提升了多词术语（MWT）的抽取效率。

From a technical perspective, terminology management systems (TMS), such as MultiTerm\textsuperscript{\textregistered}\footnote{\url{https://www.trados.com/cn/product/multiterm/}}, now support the entire lifecycle of terminology management, with interoperability facilitated by open standards such as TBX (TermBase eXchange). Recent developments feature semantic enrichment, in which termbases are linked to ontologies and knowledge graphs to support machine reasoning and dynamic modeling of legal knowledge. 
%从技术实现层面看，诸如 MultiTerm\textsuperscript{\textregistered}\footnote{\url{https://www.trados.com/cn/product/multiterm/}} 等术语管理系统（TMS）已支持术语管理的全生命周期，并通过 TBX（TermBase eXchange）等开放标准实现了系统间的互操作性。近期趋势包括语义增强，即将术语库与本体和知识图谱关联，以支持机器推理和动态法律知识建模。

In practice, termbases enable a wide range of use cases, such as translation and localization (for example, the synchronization of multilingual terminology in Swiss enterprises\footnote{\url{https://swissglobal.ch/en/services/terminology-management/}}), emergency communication, e.g., Germany's THW mobile termbase~\citep{Rsener2013TerminologiedatenbankenIM}, and legal knowledge engineering, e.g., the TERMitLEX project~\citep{peruzzo2020termitlex}. Educational initiatives have also leveraged frame semantics (such as FrameNet) to enhance training in domain-specific terminology.
%在实际应用中，术语库为多种场景提供支撑，例如翻译与本地化（如瑞士企业多语种术语同步\footnote{\url{https://swissglobal.ch/en/services/terminology-management/}}）、应急通信（如德国 THW 移动术语库~\cite{Rsener2013TerminologiedatenbankenIM}）、法律知识工程（如 TERMitLEX 项目~\cite{peruzzo2020termitlex}）。此外，教育项目还借助框架语义（如 FrameNet）提升领域术语培训的专业性。

\subsection{Cross-lingual Terminology Mapping} 
Cross-lingual terminology mapping is fundamental to overcoming barriers in multilingual legal communication. A prominent example is the InterActive Terminology for Europe (IATE) database, which has played a vital role in facilitating European Union (EU) integration by promoting consistency and precision in communication among member states \citep{vsarvcevic2016basic}. However, detailed analyses of the IATE database reveal a marked imbalance: English terms vastly outnumber those in less-represented languages such as Latvian, exposing significant disparities in multilingual coverage \citep{Karpinska_Liepina_2022}. This linguistic imbalance presents real challenges to achieving fair representation and can undermine the effectiveness of legal and administrative procedures in underrepresented languages. Similar issues can be observed in Arabic legal resources, where most existing dictionaries are limited to simple term lists and often lack contextual definitions or clear jurisdictional distinctions \citep{Halimi2024}. The predominance of English-centric resources further intensifies the shortage of legal terminology in other languages, hindering effective legal communication across jurisdictions. These disparities highlight the urgent need for robust, automated frameworks for terminology extraction, mapping, and management.
%跨语言术语映射是克服多语言法律交流壁垒的基础。在这方面，欧洲互动术语数据库（IATE）是一个具有代表性的例子。该数据库通过推动欧盟成员国之间沟通的一致性与精确性，在促进欧盟一体化进程中发挥了重要作用 \citep{vsarvcevic2016basic}。然而，对IATE数据库的深入分析表明，英语术语数量远远超过如拉脱维亚语等使用较少的语言，凸显了多语言覆盖上的显著不均衡 \citep{Karpinska_Liepina_2022}。这种语言上的不平衡对实现公平的多语言代表性构成了现实挑战，也可能削弱小语种法律与行政程序的有效性。类似的问题同样存在于阿拉伯语法律资源中，相关词典大多仅为简单的术语列表，缺乏语境定义和法域区分 \citep{Halimi2024}。以英语为中心的术语资源主导地位，进一步加剧了其他语言法律术语的匮乏，阻碍了不同法域之间的有效法律交流。这些差异凸显出亟需建立健全、自动化的术语提取、映射与管理框架。

The uneven distribution of terminological resources adds yet another layer of complexity to the problem. One practical solution, particularly when studying non-English legal systems, is to employ a well-resourced pivot language such as English or Latin \citep{Chan2011}. For example, the Rome II Regulation on the law applicable to non-contractual obligations uses Latin expressions like \textit{negotiorum gestio} and \textit{culpa in contrahendo} in its French and Italian versions to ensure clarity and uniformity across different legal traditions \citep{graziadei2025}. In East Asia, historical developments have led to many legal terms in Chinese and Japanese sharing identical or closely related sinographs (that is, classical Chinese characters or Japanese kanji). This makes certain terms function as a natural linguistic bridge. For instance, the term ``{{法人}}'' (\pinyin{\textit{fǎ rén}} in Chinese; \textit{hōjin} in Japanese) refers to ``legal entity'' in both legal systems, denoting an organization or body with its own legal personality, rights, and obligations. This shared term, written identically in both languages, offers an unambiguous bridge for legal communication and facilitates accurate cross-lingual mapping. Such shared, classical roots allow for direct correspondence, reducing ambiguity and facilitating understanding in bilingual legal contexts. In effect, using English and shared kanji as pivot elements in Chinese-Japanese legal research serves a similar bridging function to Latin in the multilingual context of the Rome II Regulation.
%术语资源分布的不均衡使相关问题愈发复杂。实际操作中，利用资源丰富的枢轴语言（如英语或拉丁语）在研究非英语法律体系时，往往能带来明显便利 \citep{Chan2011}。例如，《罗马II规约》（适用于非合同义务的法律）就在其法语和意大利语版本中引入了拉丁表达，如 \textit{negotiorum gestio} 和 \textit{culpa in contrahendo}，以保障不同法律传统下的术语清晰与统一 \citep{graziadei2025}。在东亚，由于历史发展的影响，中日两国语言中许多法律术语共享相同或相近的汉字（即古典汉字或日语汉字），因而这些术语天然成为法律交流的语言桥梁。例如，{{法人}}（中文拼音：fǎ rén，日语：hōjin）在中日法律体系中均表示``法人''，即指具有独立法律人格、权利和义务的组织或机构。该术语在两种语言中写法完全相同，含义明确，为法律交流和术语对齐提供了无歧义的桥梁。实际上，在中日法律研究中使用英语和共享汉字作为枢轴机制，正如《罗马II规约》中拉丁语在多语法律体系中的桥梁作用一样。

However, superficial similarity in Chinese characters, or in their English translations, does not guarantee that the same legal concept or meaning is being conveyed \citep{alma99156312890001452}. A telling example is the Japanese term ``{特許}'' (\textit{tokkyo}), which means ``patent,'' compared with the simplified Chinese terms ``{特许}'' (\pinyin{\textit{tè xǔ}}), meaning ``special authorization,'' and ``{专利}'' (\pinyin{\textit{zhuān lì}}), which more accurately denotes ``patent.'' While ``{特许}'' ({特别许可}) typically refers to administrative concessions or commercial franchises, ``{专利}'' is the standard term for ``patent'' in Chinese law.\footnote{The legal concept of ``patent'' itself has subtle jurisdictional differences, so even ``{特許}'' and ``{专利}'' are not perfectly equivalent.} This example illustrates that even terms with similar forms can have divergent legal meanings across jurisdictions, highlighting the complexities and potential pitfalls of legal translation and terminology mapping. Nonetheless, using a carefully selected pivot language remains an effective way to bridge linguistic and conceptual gaps in comparative legal research.
%然而，汉字表面上的相似性或其英文翻译的一致性，并不意味着相关术语在法律概念或含义上也是一致的 \citep{alma99156312890001452}。一个具有代表性的例子是日语中的``特許''（tokkyo），意为``专利''，而中文中``特许''（tè xǔ）则表示``特别授权''，``专利''（zhuān lì）才是更准确对应``patent''的术语。其中，``特许''（特别许可）通常指行政上的特许或商业特许经营，而``专利''才是中国法律中标准的``专利''术语。\footnote{``专利''这一法律概念在不同法域内也存在细微差异，因此即使``特許''和``专利''也并非完全等价。} 这个例子说明，即使是形式相似的术语，在不同法域中的法律概念也可能大相径庭，这进一步凸显了法律翻译和术语映射的复杂性与风险。尽管如此，合理选择和应用枢轴语言，仍然是比较法语境下弥合语言与概念鸿沟的有效方法。

\subsection{Historical Perspective}
The historical interplay between Chinese and Japanese legal systems has profoundly influenced the development and transmission of legal terminology in East Asia. Foundational studies show that, during Japan's modernization and China's twentieth-century legal reforms, many Japanese legal concepts and terms were extensively integrated into the Chinese legal lexicon~\citep{cho1977japanese}. Recognizing this process of historical borrowing is essential for understanding present-day challenges in legal term equivalence and translation.
%中日法律体系的历史互动极大地影响了东亚法律术语的演变与传播。基础性研究表明，在日本近代化及中国二十世纪法律变革期间，大量日本法律概念和术语被广泛引入到中国法律词汇体系中~\citep{cho1977japanese}。认识到这种历史借鉴过程，有助于厘清当下法律术语等值性与翻译难题的根源。

Comparative legal research highlights both convergences and divergences in legal terminology between the two systems. \cite{Kozanecka2018141162}, for example, employs parametric and legal-constructionist approaches to analyze how Chinese legal terms map onto those used in Japan and other East Asian jurisdictions. Illustrative cases, such as the rendering of ``不动产'' (immovable property) as ``real estate'' in the Japanese Civil Code, underscore the complexity and importance of terminology standardization. Such comparative perspectives supply both theoretical grounding and practical examples for constructing multilingual legal databases.
%比较法研究进一步揭示了中日法律术语既有趋同，也存在分歧。Kozanecka~\citep{Kozanecka2018141162} 采用参数法与法律建构方法，分析中国法律术语在日本及其他东亚法域的映射关系。具体案例如``‘不动产'译为日本民法典中的‘real estate'''等，突显了术语标准化的复杂性与必要性。这类比较视角为多语种法律数据库的构建提供了理论支撑和实际范例。

Recent years have seen major advances in legal terminology databases across East Asia. Japan's Ministry of Justice has developed the Japanese Law Translation Database System, offering official English translations of statutes and a standardized glossary~\citep{JapanLawTrans}. In China, platforms such as the National People's Congress (NPC.PRC) and PKU Law have assembled large-scale Chinese-English legal corpora and terminology banks, providing critical resources for scholars and practitioners alike. In the absence of a single central authority for terminology, corpus-driven approaches to translation and term standardization have become mainstream.
%近年来，东亚地区法律术语数据库建设取得了显著进展。日本法务省推出了``日本法律翻译数据库系统''，提供法令的官方英文译本及标准化词汇表~\citep{JapanLawTrans}。中国则有全国人大（NPC.PRC）、北大法宝等平台，构建了大规模中英法律语料库和术语库，为学界与实务界提供了重要数据资源。在缺乏统一术语权威机构的情况下，基于语料库的方法已成为翻译与术语标准化的主流。

In Chinese-Japanese legal translation, scholars agree that relying solely on character similarity or literal translation cannot achieve legal precision. More flexible strategies—blending the use of equivalent terms, paraphrasing, neologisms, and corpus-based analysis—are widely regarded as effective means to enhance consistency and quality~\citep{vsarcevic2000legal,AnSun2022,AlSaeedAbdulwahab2023}. The phenomenon of Sino-Japanese homographs is particularly noteworthy; Table~\ref{tab:term_typology} presents a typology of legal terms classified by the relationship between their written form and semantic equivalence in Chinese and Japanese law.
%在中日法律翻译实践中，学者普遍认为，仅靠汉字表面相似或直译无法实现法律意义上的精确传达。更为灵活的策略——如等值术语、释义、造词及基于语料的分析结合——被视为提升一致性与译文质量的有效路径~\citep{vsarcevic2000legal,AnSun2022,AlSaeedAbdulwahab2023}。中日同形异义（同形词）现象尤为突出，表~\ref{tab:term_typology} 按书写形式与语义等值性对法律术语进行了分类梳理。

With respect to terminology standardization, both Japan and China have converged on modern legal terms through term creation, dictionary compilation, and the establishment of standardization procedures since the Meiji era~\citep{ComparativeLegilinguistics2010,Tao2017,Qu2015}. Notably, cross-national initiatives such as the Nagoya University Legal Information Project are advancing automated term extraction, alignment, and keyword-in-context search technologies for East Asian legal systems\footnote{\url{https://jalii.law.nagoya-u.ac.jp/enproject}}. Although a fully comprehensive Chinese-Japanese legal terminology database is still in progress, the growing accumulation of resources and collaborative platforms lays a strong foundation for future standardization and legal interoperability.
%在术语标准化方面，自明治维新以来，中日两国通过术语创制、词典编纂与标准化程序逐步实现了现代法律术语的趋同~\citep{ComparativeLegilinguistics2010,Tao2017,Qu2015}。诸如名古屋大学法律信息项目等跨国协作，正在推动东亚法域自动化术语抽取、对齐与语境检索等关键技术的发展\footnote{\url{https://jalii.law.nagoya-u.ac.jp/enproject}}。虽然目前尚未建立完善的中日法律术语数据库，但相关资源与合作平台的不断积累，为未来标准化和法律互操作性奠定了坚实基础。

\begin{table}[ht]
\centering
\caption{Typology of Chinese-Japanese legal terms: form and meaning correspondence.}\label{tab:term_typology}
\begin{tabular}{p{3.5cm} p{2cm} p{2cm} p{4cm}}
\toprule
\textbf{Type} & \textbf{Chinese Term} & \textbf{Japanese Term} & \textbf{Explanation / Example} \\
\toprule
Identical Form and (Nearly) Identical Meaning
  & 监护 (jiānhù) 
  & 監護 (kango) 
  & ``Guardianship''; civil law concept and general institutional logic highly similar, though procedural details may differ. \\
\midrule
Identical Form, Different Meaning (False Friends)
  & 裁判 (cáipàn) 
  & 裁判 (saiban) 
  & Chinese: judgment/decision/referee (broad); Japanese: strictly ``judicial trial/court judgment''. \\
\midrule
Different Form, Same/Synonymous Meaning
  & 合同 (hétóng) 
  & 契約 (keiyaku) 
  & ``Contract''; core civil law meaning aligned, but legal traditions and doctrinal boundaries can diverge. \\
\midrule
Partial Overlap
  & 法人 (fǎrén) 
  & 法人 (hōjin) 
  & ``Legal entity''; general principle similar, but entity types and registration systems are not always identical. \\
\midrule
System-specific Term
  & 土地承包经营权 (tǔdì chéngbāo jīngyíngquán) 
  & --- 
  & Exists only in Chinese rural land law; no direct Japanese equivalent. \\
\bottomrule
\end{tabular}
\end{table}

\subsection{Terminology Extraction and Alignment} 
For monolingual terminology extraction, statistical measures such as TF-IDF and TextRank are typically used to identify candidate terms within each language. This initial selection is then followed by expert legal validation to ensure doctrinal accuracy~\citep{manning1999foundations}. The validated terms are mapped to legal concept nodes and further enriched with metadata, including jurisdiction, enactment dates, hierarchical relations, and cross-lingual links. This structured, often graph-based representation facilitates efficient retrieval, faceted search, and robust version control.
%在单语法律术语抽取中，常用的统计方法包括TF-IDF和TextRank等，用于在各自语言中识别候选术语，随后由法律专家进行人工审核，以确保符合教义学要求~\citep{manning1999foundations}。最终通过的术语被映射到法律概念节点，并进一步补充法域、颁布时间、层级关系和跨语链接等元数据。这种结构化、通常采用图模型的表示方式，有助于高效检索、分面化查询和强大的版本控制。

Recent advances in artificial intelligence have substantially enhanced each step of this pipeline. Techniques such as word embeddings, BERT-style encoders, and legal knowledge graphs are now used to cluster synonyms and uncover hidden term variants~\citep{ghanem2023benchmark}. Large language models (LLMs)—including GPT-4, Llama-3, and DeepSeek-v3~\citep{openai2023gpt4, touvron2023llama, deepseekai2025deepseekv3technicalreport}—can generate candidate definitions, suggest cross-jurisdictional matches, and flag translation inconsistencies. Retrieval-augmented generation (RAG) methods further reinforce terminological consistency in downstream tasks such as statute summarization and legal machine translation~\citep{lewis2020retrieval, NEURIPS2024_6ddc001d}. Generative systems like ChatGPT have demonstrated the capacity to automatically draft nearly every dictionary component, from corpus-derived entry skeletons to fully polished, structured definitions~\citep{deschryver2023generative, li2024using}.
%近年来，人工智能的进步极大提升了术语抽取流程的各个环节。词向量、BERT类编码器和法律知识图谱等技术可聚类同义词并发现隐性变体~\citep{ghanem2023benchmark}。大型语言模型（如GPT-4、Llama-3、DeepSeek-v3）\citep{openai2023gpt4, touvron2023llama, deepseekai2025deepseekv3technicalreport} 能自动生成术语定义、建议跨法域对应，并标注翻译不一致项。检索增强生成（RAG）方法进一步提升法规摘要和法律机翻等下游任务的术语一致性\citep{lewis2020retrieval, NEURIPS2024_6ddc001d}。以ChatGPT为代表的生成式系统已能自动起草从语料抽取到结构化定义等几乎全部词典组件~\citep{deschryver2023generative, li2024using}。

Despite these technological advances, significant challenges remain in legal NLP~\citep{ariai2024legal_review}. Recent reviews emphasize persistent issues with named entity recognition, term boundary detection, data sparsity, and model interpretability—particularly with respect to segmentation and alignment in multilingual legal contexts.
%尽管技术不断进步，法律NLP领域仍面临诸多难题~\citep{ariai2024legal_review}。最新综述指出，在实体识别、术语边界判定、数据稀疏和模型可解释性等方面，尤其是在多语法律分词和对齐任务中，问题尤为突出。

Prior approaches to multilingual legal terminology extraction and alignment can be broadly categorized into three types. The first, Statistical Machine Translation (SMT) and phrase-based models, relies on co-occurrence statistics and surface alignments. These methods often struggle to accurately identify legal terms in Chinese and Japanese, where the lack of explicit word boundaries leads to fragmented or ungrammatical extractions~\citep{koehn-knowles-2017-six, kubota2002mostly, zhang-etal-2006-subword}. Such models typically lack the precision required for legal-domain applications. The second, Transformer-based Neural Machine Translation (NMT), has improved general translation quality but still faces challenges with long, complex legal sentences and domain adaptation; limitations in the attention mechanism can result in unstable alignments and misplaced legal terms~\citep{koehn-knowles-2017-six, zhuocheng-etal-2023-scaling}. These models often do not reliably capture the legal semantics needed for high-quality term extraction. The third, recent LLM-based approaches (e.g., GPT-4, Mixtral), have shown improved F1 scores in legal terminology extraction, but continue to encounter difficulties with precise term boundaries, contextual adaptation, and the lack of end-to-end quality assurance or systematic update mechanisms~\citep{breton2025llm_lex_extraction}.
%以往多语法律术语抽取与对齐的方法大致可分为三类。第一类为统计机器翻译（SMT）和短语模型，依赖共现统计和表层对齐，但因中日语缺乏显式分词，常导致术语碎片化或语法错误~\citep{koehn-knowles-2017-six, kubota2002mostly, zhang-etal-2006-subword}，难以满足法律领域精度需求。第二类为基于Transformer的神经机器翻译（NMT），虽提升了通用翻译质量，但在处理长句、复杂法律结构和领域适应性方面仍有不足，注意力机制的限制会造成术语对齐不稳定或定位错误~\citep{koehn-knowles-2017-six, zhuocheng-etal-2023-scaling}，难以实现高质量术语抽取。第三类为新近的LLM方法（如GPT-4、Mixtral），在术语抽取F1分数上取得进展，但仍面临边界判定、语境适应和缺乏端到端质量保障与系统性更新机制等问题~\citep{breton2025llm_lex_extraction}。

\section{Methodology}
This section presents the methodology developed to address the research questions outlined above. To tackle the persistent challenges of cross-lingual legal terminology mapping - especially for resource-scarce language pairs like Chinese and Japanese, we propose a human-AI collaborative workflow built on a multi-agent system. This system integrates large language models (LLMs) and domain experts to automate and validate each key step, including OCR, article-level alignment, terminology extraction, and multidimensional quality assurance. English serves as a semantic bridge, enhancing both the accuracy and disambiguation of terminology alignment between Chinese and Japanese. In addition, a few-shot learning is applied to mitigate data scarcity in low-resourced legal subdomains. 
%本节介绍为解决前述研究问题而设计的方法论。针对跨语言法律术语映射的持续挑战，尤其是中、日等资源稀缺语种对，本研究提出了一种人机协同的多智能体流程。该系统融合了大语言模型（LLMs）与领域专家，自动化并验证包括OCR、条款级对齐、术语抽取及多维质量评估在内的关键环节。英语作为语义桥梁，有效提升了中日术语对齐的准确性与消歧能力。

Traditional legal translation methods drawing on bilingual dictionaries, professional translators, and a relatively small set of parallel legal texts often prove inadequate for the demands of today's legal communication landscape. In our experience, these tools may work reasonably well for routine documents, but when it comes to handling large volumes of legal material or dealing with the intricate concepts and specialized language found in statutory or regulatory texts, their limitations quickly become apparent. Standard machine translation systems, for their part, rarely succeed in capturing the subtlety and specificity of legal terminology; domain terms are frequently mistranslated or flattened into vague generalities.

Adding to this problem is a notable imbalance in multilingual legal resources. English remains the dominant pivot, while resources for many other languages, especially those less represented in global legal discourse, are often incomplete or entirely lacking. For researchers and practitioners working in these contexts, the absence of comprehensive and reliable term mappings can lead to inconsistency, loss of nuance, and ultimately legal misunderstandings. It has become increasingly clear that new, more automated approaches to legal terminology mapping are needed: approaches that do not just translate words but can account for conceptual distinctions and preserve the intended meaning across legal systems.

\begin{figure}[!htbp]
\centering
\includegraphics[width=\textwidth]{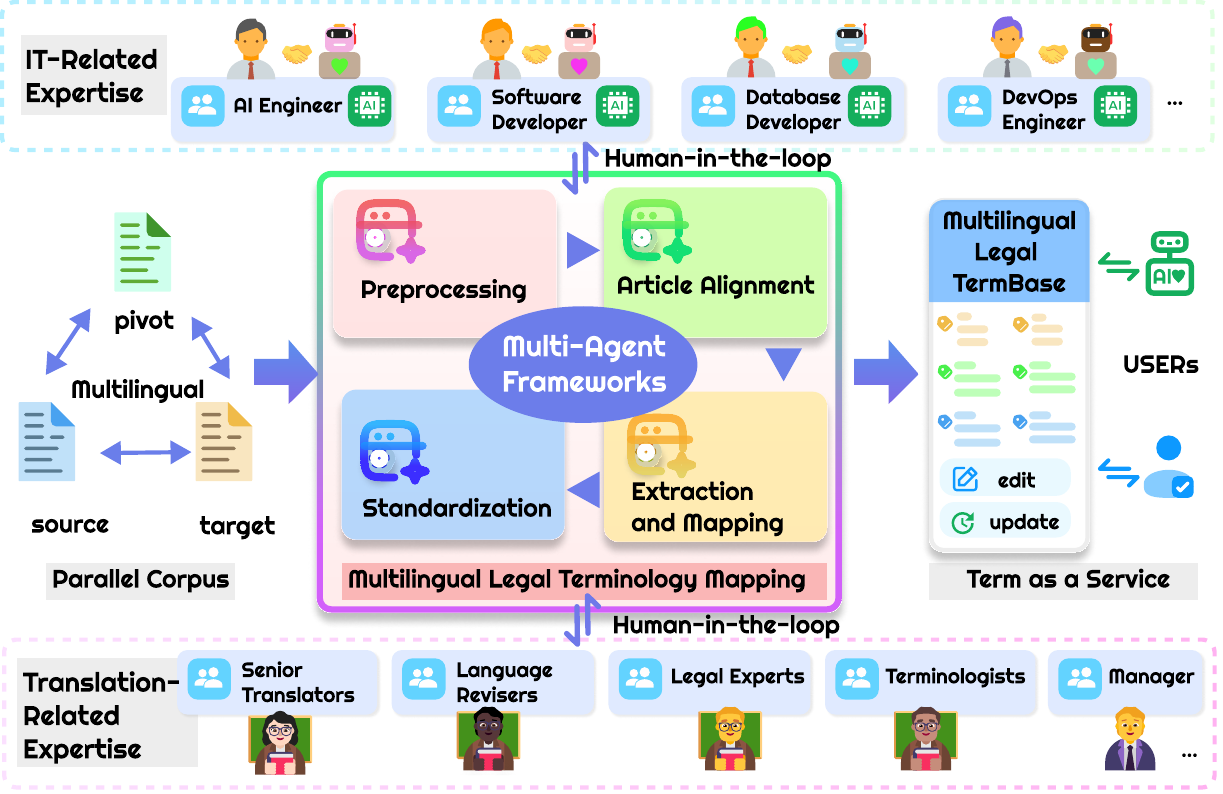}
\caption{Overview of our multi-agent framework for Multilingual Legal TermBase Construction with human-AI collaboration. All external human experts can be excluded from the process, as the terminology extraction framework is capable of running autonomously with the multi-agent system. However, the involvement of human experts further enhances the accuracy and reliability of the extraction results.}\label{scheme}
\end{figure}

In response to these challenges, our team developed a workflow that brings together human expertise and AI-based tools in a genuinely collaborative fashion. Figure~\ref{scheme} illustrates how to complete multilingual legal terminology mapping using a multi-agent framework
with human-in-the-loop. Rather than relying on assumptions or generic templates, we put our methodology to the test on a challenging trilingual dataset: 35 core Chinese statutes and their English and Japanese translations. The results were revealing. By comparing our approach to traditional manual methods and standard automated tools, we found consistent improvements in terminology coverage, accuracy, and scalability, particularly in areas where prior resources were thin or inconsistent.

At the core of the framework is a multi-agent system that integrates the precision of lexicographic standards, the efficiency of advanced AI automation, and the agility of a continuous delivery pipeline. This synergy transforms traditional static legal dictionaries into dynamic, multilingual terminology resources, expanding conceptual coverage, improving semantic granularity, and allowing rapid adaptation to legal and linguistic change. Such qualities are indispensable for reliable cross-jurisdictional legal communication in today's interconnected world.
%本框架的核心是多智能体系统，融合了词典学的严谨标准、先进AI自动化的高效处理能力以及持续交付机制的灵活性。这种协同不仅实现了传统静态法律词典向动态多语术语资源的转型，还大幅提升了术语库的概念覆盖、语义颗粒度和对法律与语言变迁的快速响应能力，为当前全球化语境下的跨法域法律交流提供了坚实保障。

To enable scalability and sustainable development, we adopt a cloud-based ``{Terminology-As-A-Service}'' ({TAAS}) architecture. The platform supports collaborative editing dashboards, CI/CD pipelines for seamless term updates, and granular access controls, allowing real-time entry refinement with authoritative oversight. 
%为保障系统的可扩展性与可持续发展，我们采用了云端``术语即服务''（TAAS）架构。该平台支持协作编辑面板、CI/CD自动化术语更新流程与细粒度权限控制，实现条目的实时完善与权威审核。针对低资源法律子领域，还引入了少样本学习以缓解数据稀缺。
  
To fulfill these requirements, our workflow orchestrates a suite of specialized LLM-based agents, including multimodal models such as GPT-4.1, Gemini-2.5, Claude-4, DeepSeek-v3 and Qwen3 serials. The full pipeline comprises five main stages: (1) data collection and preprocessing, (2) bilingual (even trilingual) sentence alignment, and (3) terminology extraction and mapping, (4) terminology standardization and (5) systematic evaluation and quality assurance. Once the conceptual backbone is established, editors generate precise bilingual and trilingual term equivalents, add authoritative references, and annotate usage constraints. Iterative review by domain experts and pilot users ensures clarity, consistency, and legal reliability before the Multilingual Legal Terminology Database (MLTDB) is made available via both human-friendly interfaces and programmatic APIs. The system further supports ongoing updates in response to legislative amendments or landmark legal decisions, guaranteeing the resource remains current and authoritative. 
%为此，我们提出的工作流程整合了多种基于LLM的智能体（如GPT-4o-mini、GPT-4o、DeepSeek等多模态模型），涵盖三大阶段：(1) 语料预处理，(2) 双语/三语句对对齐，(3) 术语抽取与质量评估。当术语概念骨架搭建完成后，编辑团队会为每个概念拟定精确的三语对应术语，补充权威出处与使用约束，经领域专家和试点用户多轮审核后，术语库正式上线并同时支持人机接口与程序化API访问。系统还设有持续维护机制，可根据法律修订或重大判例动态更新，确保术语资源的时效性与权威性。

Throughout the entire life-cycle of legal termbase construction, we ensure that at least two or three human experts are involved at every stage. These experts include Senior Translators, Language Revisers (quality assurance), Legal Experts, Terminologists, Software Developers, Database Developer, AI Engineers, DevOps 
Engineer, and Managers. Experts are actively involved in overseeing each step of the process, from initial term extraction and alignment to the final review and validation of the termbase. Their roles include manual intervention when necessary, providing expertise in legal nuances, linguistic accuracy, and terminological consistency, as well as ensuring the applicability of terms in legal contexts.
%在整个法律术语库构建生命周期过程中，我们确保在每个阶段至少有两到三位专家参与。这些专家包括资深翻译、语言修订员、法律专家、术语学专家、软件开发人员、AI工程师、运维工程师和项目经理。专家们积极参与监督每个步骤，从初步术语提取和对齐到最终术语库的审核与验证。他们的职责包括在必要时进行人工干预，提供法律细节、语言准确性和术语一致性的专业意见，并确保术语在法律语境中的适用性。

As shown in Figure~\ref{scheme}, the experts collaborate closely with the AI models, guiding them where required and performing manual validation to ensure the quality of the extracted terminology. The final review is conducted by the experts, ensuring that the terms align with legal standards, linguistic norms, and domain-specific requirements. These rigorous quality control measures, which integrate both AI and expert oversight, guarantee the precision, consistency, and legal applicability of the term definitions, ensuring that the termbase meets the highest standards of quality and reliability. In the following four subsections the main stages of the approach are explained in detail, including preprocessing, article alignment, extraction and mapping, and standardization.
%专家们与AI模型密切合作，在需要时提供指导，并对提取的术语进行人工验证，确保其质量。最终审核由专家完成，确保术语符合法律标准、语言规范和特定领域要求。这些严格的质量控制措施，结合AI与专家的监督，确保了术语定义的精确性、一致性和法律适用性，确保术语库达到最高的质量和可靠性标准。
\subsection{Preprocessing}

The process begins with careful planning. This stage requires the research team to fix the scope of legal systems (Chinese, Japanese, and English), profile the future user groups (practitioners, translators, scholars, and NLP systems), and negotiate the depth of information for each entry. The team also establishes a detailed update schedule to accommodate the continuous evolution of statutes, case law, and administrative regulations. Once the blueprint is in place, we gather a large, balanced corpus of legal texts. We select 35 current Chinese legal statutes enacted or amended during 30 years between 1995 and 2025 (including 1 civil code, 34 full laws) from the National Legal Regulations Database\footnote{\url{https://flk.npc.gov.cn/index.html}}, along with their English translation from the public official websites such as the National People's Congress official website\footnote{\url{https://english.www.gov.cn/archive/lawsregulations}}, as well as Japanese translations mainly from the Japan External Trade Organization (JETRO)\footnote{\url{https://www.jetro.go.jp/world/asia/cn/ip/law/}}. These texts cover a wide range of fields, including the Constitution, administrative law, civil and commercial law, and social law, ensuring the authority and applicability of the terminology. Table~\ref{table1} shows the statistics of built parallel corpus in each language. The detailed information of legal categories and names of the laws included in the Chinese-Japanese-English legal corpus are listed in Appendix~\ref{secB}. 

\begin{figure}[t]
\centering
\includegraphics[width=\textwidth]{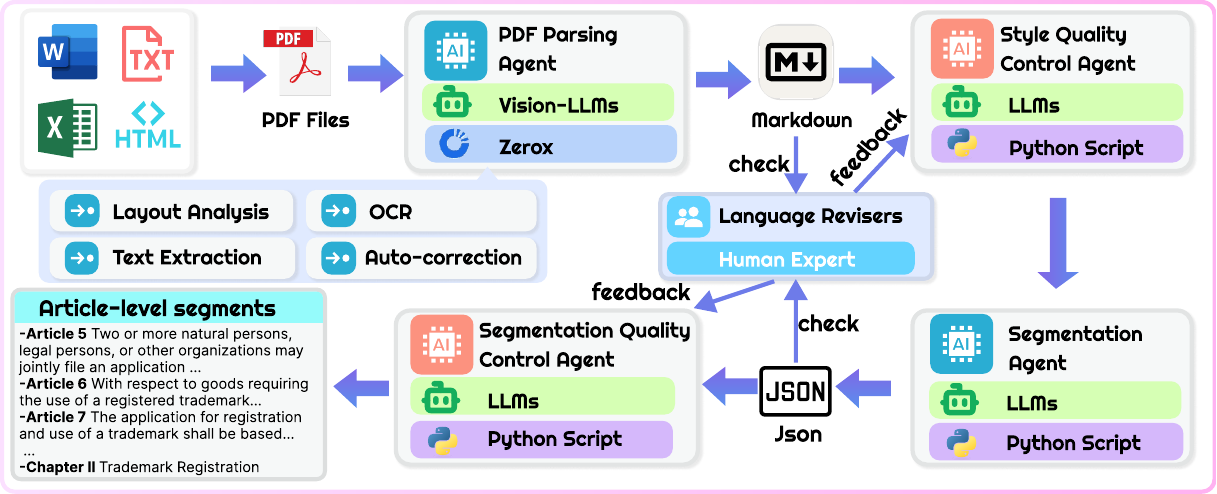}
\centering\caption{Multi-agent workflow for legal document preprocessing to generate article-level segments.}\label{preprocess-workflow}
\end{figure}

Figure~\ref{preprocess-workflow} shows how to generate article segments using a multi-agent workflow. To efficiently process various types of original legal documents, such as scanned images, PDFs, or Word files, we first manually convert all documents into a unified PDF format. Subsequently, four intelligent agents are constructed: the PDF Parsing Agent, the Style Quality Control (also called Quality Assurance) Agent, the Content Segmentation Agent, and the Segmentation Quality Control Agent. These agents collaborate to transform PDFs into structured plain-text corpora segmented at the article level.

The PDF Parsing Agent is built on top of the Zerox\footnote{\url{https://github.com/getomni-ai/zerox}} software. To reduce computational cost, we employ the GPT-4.1-mini multimodal model as the OCR engine, which accurately identifies document structure and extracts paragraphs in Chinese, Japanese, and English, enabling efficient and clean text extraction. The Style Quality Control Agent, powered by the more capable LLMs such as GPT, Gemini, DeepSeek, executes contextual instructions to perform comprehensive text cleaning (e.g., removing special characters, blank lines, headers, footers, page numbers, irrelevant URLs, and annotations), paragraph reordering (based on logical structure), and automatic correction (e.g., spelling and grammatical errors), generating well-formatted intermediate Markdown documents. The Content Segmentation Agent processes these intermediate documents using a hybrid approach: initially applying rule-based Python scripts (e.g., regular expressions and pattern matching) to conduct coarse segmentation, followed by leveraging DeepSeek-v3's language understanding capabilities to further segment texts into logical units such as articles, chapters, and sections. The resulting structured content is stored in JSON format. The Segmentation Quality Control Agent then reviews the segmented results using the more advanced DeepSeek-r1 model to reprocess any errors or overly long text blocks.

Language revisers are employed to review whether the segmented results meet the required standards after the PDF parsing and statute segmentation steps. These process involves manual inspection by experts, who assess both the formatting and the linguistic quality of the segmented text. The reviewers then provide targeted suggestions to the Style Quality Control Agent and the Segment Quality Control Agent, enabling these intelligent agents to refine and correct the output accordingly. This hybrid approach ensures that both the stylistic and structural aspects of the segmented statutes adhere to professional and domain-specific requirements. Human experts supervise and validate the intermediate Markdown and JSON outputs, providing corrective feedback to the agents and optimizing final results through human-in-the-loop refinement. 
\begin{table}[!htbp]
\centering
\caption{Length distribution statistics of article-level segments.}
\begin{tabular}{>{\raggedright\arraybackslash}m{1.5cm}>{\centering\arraybackslash}m{1cm}>{\centering\arraybackslash}m{1.8cm}>{\centering\arraybackslash}m{1cm}>{\centering\arraybackslash}m{2cm}>{\centering\arraybackslash}m{1.5cm}}
\toprule
\textbf{Language} & \textbf{Entries} & \textbf{Avg Words} & \textbf{$\pm$std.} & \textbf{Total Words} & \textbf{Ratio} \\
\toprule
Chinese & 5,172 & 42.0 & $\pm$37.5 & 249,405 & 24.4\%  \\
English & 5,172 & 70.0 &  $\pm$60.5 & 367,863 & 36.1\% \\
Japanese & 5,172 & 75.1 &  $\pm$66.2 & 403,525 & 39.5\%  \\
\bottomrule
\end{tabular}\label{table1}
\end{table}

\subsection{Article Alignment}
\begin{figure}[t]
\centering
\includegraphics[width=\textwidth]{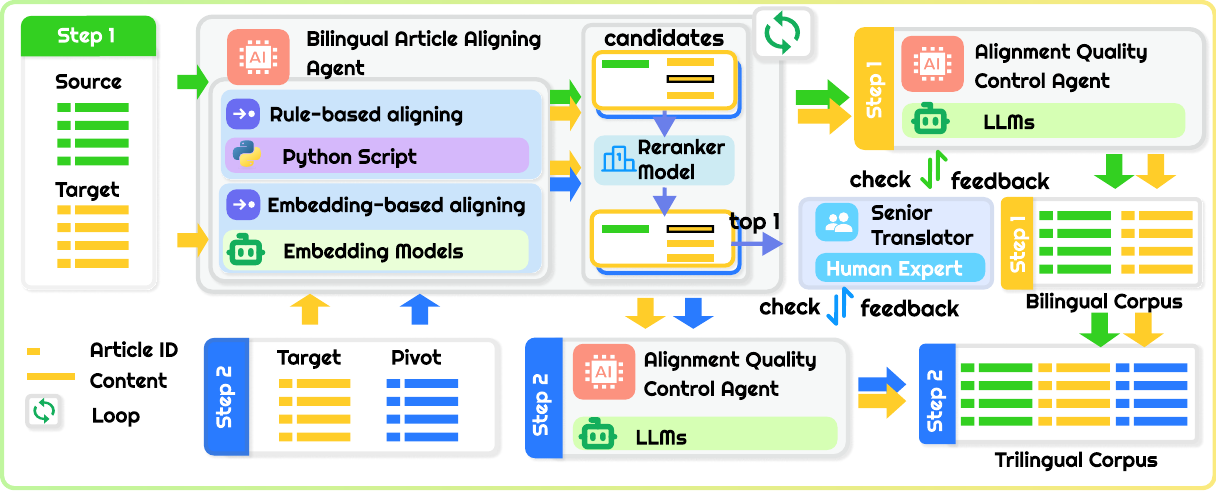}
\centering\caption{Multi-agent workflow for article alignment and multilingual parallel corpus construction.}\label{alignment}
\end{figure}

Figure~\ref{alignment} presents the process of achieving article-level alignment across the source, target, and pivot languages. In the first step, the Bilingual Article Aligning Agent employs a mixed strategy combining rule-based alignment using a Python script and embedding-based alignment with the OpenAI Text Embedding model\footnote{\url{https://platform.openai.com/docs/models/text-embedding-3-small}}
. The input consists of source and target legal texts (e.g., Chinese and Japanese) at the article level. The agent aligns these articles based on predefined rules and semantic embeddings. For example, a predefined rule might map terms like ``第X条'' or ``第X章'' in Chinese and Japanese to ``Article'' or ``Chapter'' in English legal texts. The embedding-based alignment computes cosine similarity using embeddings generated by the OpenAI text embedding model. Given a source article, alignment candidates are generated by merging the outputs from the rule-based aligning and embedding-based aligning modules. Then, we employ reranker models like Jina\footnote{\url{https://huggingface.co/jinaai/jina-reranker-v2-base-multilingual}}
 or BGE-m3\footnote{\url{https://huggingface.co/BAAI/bge-reranker-v2-m3}}
 to re-score the candidates and output the best aligned article in the target or pivot language. If the alignment passes the examination, it proceeds to the next step. In the event of alignment failure, an automatic retry mechanism is triggered, and the failure feedback is sent to the article aligning agent for re-running, ensuring that the alignment process ultimately achieves a 100\% success rate. The Alignment Quality Control Agent examines the alignment between the source and target/pivot languages. This agent primarily provides quality assessment suggestions to human experts, who make the final decision on whether the aligned content should be stored in the bilingual parallel corpus or require re-alignment.

After achieving bilingual article-level alignment, this process continues to add the third (i.e., pivot) language like English, yields a trilingual corpus. This multilingual corpus allows for better cross-lingual legal comparisons and enhances the ability to perform more precise legal term mapping across multiple legal systems. In essence, this system automates and ensures high-quality alignment between legal texts at the article level. The integration of advanced embedding techniques and quality control measures ensures both accuracy and scalability in legal text alignment and corpus development.

Finally, a senior translator is embeded as a reviewer to examine and validate all alignment results. This expert ensures the quality and reliability of the parallel corpus and supervises the alignment quality control agents, providing additional guidance (e.g., prompt) and oversight to further enhance their performance. These texts were also annotated by legal domain, promulgation date, and language, providing a clean and controllable corpus foundation for subsequent terminology extraction and analysis. Since this section involves numerous engineering optimizations and practical know-how, we will not elaborate further.

\begin{table}[!htbp]
\centering
\caption{Prompt for zero-shot and few-shot used to perform bilingual legal term extraction (e.g., Chinese-English).}\label{prompt-table}
\begin{tabular}{p{12.5cm}}
\toprule
\textbf{Instruction}: You are a professional bilingual legal terminology extraction expert, especially for Chinese and English. Your task is to accurately identify and extract professional term pairs from the provided bilingual legal text data. \\
\midrule
\textbf{Input Format}: The input format is: [Chinese text] \texttt{\textbackslash t} [English text]\\
\midrule
\textbf{Requirements}:
Please strictly follow the processing rules below: \\
1. \textbf{Term Extraction}: Extract all professional term pairs from a single input and output them as a JSON array. \\
2. \textbf{Semantic Correspondence}: Ensure that the Chinese and English terms correspond completely in the professional context. \\
3. \textbf{Context Accuracy}: The ``context'' field must directly quote the original Chinese sentence fragment.\\
4. \textbf{Intelligent Explanation}: Add concise explanations for terms that are highly specialized or ambiguous. \\
5. \textbf{Format Specifications}: Output pure JSON array content, no JSON tags, no additional text or labels.  \\
\midrule
\textbf{Output Format}: \\
\{``terms'':[ \\
 \quad \quad\quad\quad\quad\quad\{ ``chinese'': ``source term'', \\
 \quad \quad\quad\quad\quad\quad\quad``english'': ``target term'', 
 \\
 \quad \quad\quad\quad\quad\quad\quad``context'': ``source sentence fragment'',  \\ \quad\quad\quad\quad\quad\quad\quad``en\_context'': ``target translation fragment'', \\
\quad\quad\quad\quad\quad\quad\quad``explanation'': ``explanation in Chinese''  \}, \\
\quad\quad\quad\quad\quad\quad\quad \dots // \texttt{other terms} ] \\
\} \\
\midrule
\textbf{Examples}: // \texttt{optional, not required for zero-shot}\\
\textbf{Example 1}: \{ \text{Input} + \text{Output} 
\}, \textbf{Example 2}: \{ \text{Input} + \text{Output}\}, \dots \\
\midrule
\textbf{User Input}:  
企业研制新产品、改进产品，进行技术改造，应当符合本法规定的标准化要求。 \texttt{\textbackslash t} Where enterprises improve their products, develop new products, or upgrade technology, they shall meet standardization requirements as stipulated in this Law. // \texttt{a real example}\\
\bottomrule
\end{tabular}
\end{table} 
\subsection{Extraction and Mapping}

\begin{figure}[!htbp]
\centering
\includegraphics[width=\textwidth]{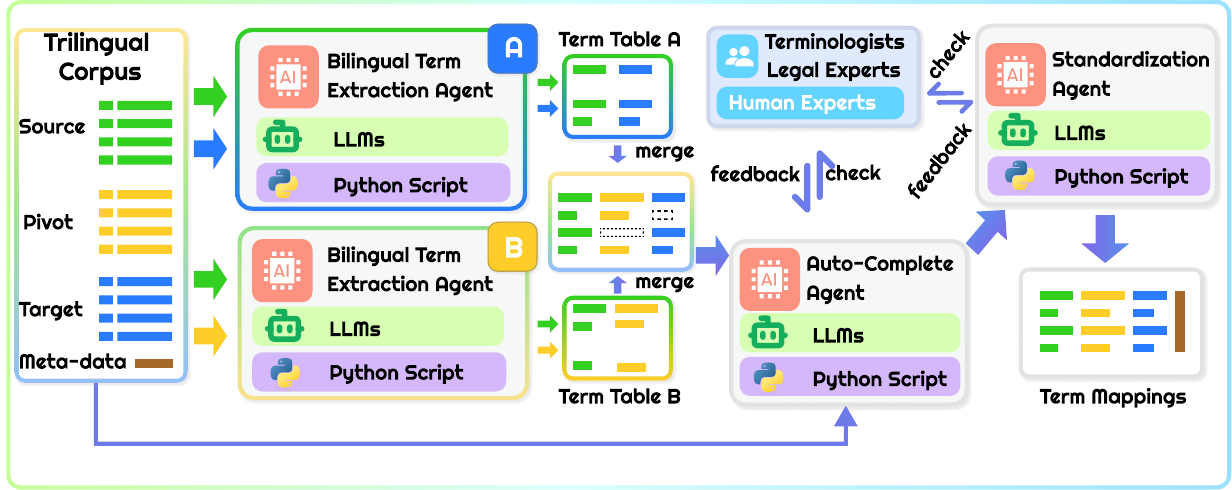}
\caption{Terminology extraction and mapping along with standardization. The dual-stream extraction approach using two bilingual term extraction branch to get better term mappings.}\label{mapping}
\end{figure}

Figure~\ref{mapping} shows the details of terminology extraction and mapping. After generating high-quality, aligned trilingual article triplets, this study introduces three specialized intelligent agents: the Bilingual Term Extraction Agent, the Auto-Complete Agent, and the Term Standardization Agent.
Distinct from traditional term extraction methods that primarily rely on frequency statistics or neural sequence modeling, our approach leverages the advanced language understanding capabilities of large language models (LLMs). The Bilingual Term Extraction Agent autonomously extracts key information from each aligned legal article set, including Chinese and Japanese legal terms, their English equivalents, source citations, relevant legal articles, context in the original Chinese text, translations of English and Japanese, and explanation.

To enhance extraction robustness and accuracy, we propose a dual-stream extraction strategy. Table~\ref{prompt-table} details the specific bilingual term extraction prompts utilized by the key agents. Each final input prompt is structured as ``task instruction + input format + additional requirements + output format + examples (optional).''  To address the issue of missing English or Japanese terms in trilingual legal terminology entries extracted from bilingual corpora, we have developed an intelligent Auto-complete Agent. This agent automatically identifies entries where either the English or Japanese term is absent and suggests accurate completions based on the available Chinese term and its contextual information. Leveraging state-of-the-art large language models and aligned legal corpora, the system analyzes the semantic and legal context of each entry to generate high-quality term suggestions. This approach not only streamlines the terminology curation process but also enhances the coverage, consistency, and usability of the multilingual legal termbase.

\subsection{Standardization}
To ensure the quality and consistency of our multilingual legal terminology resource, we implemented a systematic terminology standardization process. For each Chinese legal term, all extracted translation variants were evaluated according to a clear set of criteria including translation accuracy, professionalism, standardization, context quality, and fluency. Table~\ref{tab:standardization_workflow} summarizes the evaluation and selection process.

\begin{table}[!htbp]
\centering
\caption{Prompt for multilingual legal terminology standardization.}
\label{tab:standardization_workflow}
\begin{tabular}{p{12.5cm}}
\toprule
\textbf{Instruction}: You are a senior legal terminology expert, fluent in Chinese, Japanese, and English.\\
\toprule
\textbf{Tasks}: Your tasks are:\newline
1. To evaluate the quality of different translation variants.\newline
2. To select the best translation. \newline
3. To merge similar translations, but preserve at least 3 variants with different meanings.\newline
4. To pay special attention to the accurate translation of proper nouns and specific legal terms.\\
\midrule
\textbf{Criteria}:  \\
\textbf{-Translation accuracy}: whether the translation accurately reflects the meaning of the Chinese term.\newline
\textbf{-Professionalism}: whether the correct legal terminology is used in the target language.\newline
\textbf{-Standardization}: whether the translation follows standard conventions (e.g., capitalization, singular/plural forms).\newline
\textbf{-Context quality}: whether the context information is rich and accurate. \newline
\textbf{-Fluency}: whether the translation is natural and idiomatic in the target language.\\
\midrule
\textbf{Important Constraints:}:  \\
$\cdot$You may only choose from the variants provided below and may not generate new content.\newline
$\cdot$You must not modify, combine, or create new translations.\newline
$\cdot$All content must strictly come from the existing variants.\\
\midrule
\textbf{Output Format}: $\dots$ \texttt{//omitted for brevity.} \\
\bottomrule
\end{tabular}
\end{table} 
All translation variants and their contextual examples were reviewed by a Standardization Agent. In practice, there are two distinct cases handled by the Standardization Agent:
\begin{enumerate}
\item When there is a clear, best variant that accurately represents the term's meaning, the agent selects that variant and disregards others. In this case, no modification, merging, or creation of new translations occurs.
\item In cases where variants are semantically similar but differ in certain minor aspects (e.g., singular vs. plural forms, or small grammatical differences such as the inclusion of ``the'' vs. ``a,'' or ``have'' vs. ``has''), the agent merges the variants based on grammatical rules to form a standardized entry that maintains the intended meaning while ensuring consistency in form.
\end{enumerate}
As a result, the best variant was selected, otherwise，where appropriate, elements from multiple variants were merged to form a standardized entry. In cases where variants represented substantially different meanings, they were preserved as distinct entries. 

Besides, legal experts also play a critical role primarily in the term mapping and standardization stages. Their involvement ensures that the mapped term pairs across languages are both semantically precise and contextually appropriate. During this phase, legal experts review and validate the candidate term mappings proposed by the Term Mapping Agent, resolve any ambiguities or inconsistencies, and provide authoritative input for the final standardization of terminology, resulting multilingual terminology database adheres to domain-specific standards and legal accuracy.

\section{Experimental Results}\label{sec2}
To evaluate the effectiveness of our agent-based workflow for legal terminology extraction, as discussed in above sections, we conducted experiments using multiple large language models (LLMs) to process a subset of the whole dataset: the Trade Union Law of the People's Republic of China (2001), Standardization Law (2017) and Against Unfair Competition Law (2019). These three laws were selected for their representation of distinct legal domains, providing a diverse set of terminological challenges for the models. For the experiments, we focused on a multilingual approach, incorporating Chinese, Japanese, and English, rather than limiting the analysis to just bilingualism, allowing for a more comprehensive evaluation of the models' ability to handle cross-lingual legal terminologies.

\subsection{Effectiveness of Dual-stream Extraction}
Table~\ref{tab:extractor_comparison} presents a comprehensive comparison of extraction results under different experimental settings for the two laws, the Standardization Law and Trade Union Law. Each row corresponds to a unique configuration defined by the combination of the underlying extraction method (dual extractor vs. single extractor, \textbf{Dual}), prompt setting (few-shot vs. zero-shot), and completion strategy (auto-complete vs. no-complete, \textbf{Auto-Compl. }). The table reports key metrics including success rate (\textbf{Success Rate})\footnote{The ``success rate'' refers to the coverage metric, which measures the proportion of articles from the total set of law articles that successfully yield at least one term mapping. This metric reflects the ability of the system to generate term mappings for a given article, irrespective of the correctness of the extracted terms.}, total (\textbf{Extracted}) and unique terms extracted (\textbf{Unique}), average terms per article (\textbf{Avg}), and the coverage of Japanese (\textbf{JA}) and English (\textbf{EN}) terms.
\begin{table}[!htbp]
\centering
\caption{Performance comparison of terminology extraction and mapping using universal approach (DeepSeek-v3) \textit{w/} and \textit{w/o} dual-stream approach (\textbf{Dual}) and auto-complete agent (\textbf{Auto-Compl.}) on the Standardization Law and the Trade Union Law. }\label{tab:extractor_comparison}
\begin{tabular}{c|l|ccrrrrrr} 
\toprule
&  & \textbf{Dual}  & \textbf{Auto-Compl.} &  \textbf{Succ. Rate} & \textbf{Extracted} & \textbf{Unique} & \textbf{Avg}  & \textbf{JA} & \textbf{EN}  \\
\midrule
\multirow{9}{*}{\rotatebox{90}{\textbf{Standardization}}} & \multirow{4}{*}{\rotatebox{90}{\textbf{few-shot}}}& \cellcolor{lightblue}\checkmark & \cellcolor{lightblue}\checkmark       &  \cellcolor{lightblue}100.0\% & \cellcolor{lightblue}356 & \cellcolor{lightblue}353 & \cellcolor{lightblue}6.6 & \cellcolor{lightblue}353 & \cellcolor{lightblue}353 \\
&   & \checkmark &  &  100.0\% & 345 &342	&6.4  &286  &  288\\
&   &  & \checkmark &	100.0\%	&246	& 245	&4.6  &245&	245\\ 
&  & & &    85.2\%	& 200 & 	198& 	4.3	& 198	& 198\\ \cmidrule{2-10}
& \multirow{4}{*}{\rotatebox{90}{\textbf{zero-shot}}} &  \cellcolor{lightred} \checkmark &  \cellcolor{lightred}\checkmark 	&\cellcolor{lightred}100.0\%&	\cellcolor{lightred}351	&\cellcolor{lightred}349&	\cellcolor{lightred}6.5	&\cellcolor{lightred}349&	\cellcolor{lightred}349\\
& &  \checkmark 	& &100.0\% &	311	 &310 &	5.8 &	260	 &254\\
 &  &  & \checkmark 	& 100.0\% &	253	 &249 & 4.7	&   249	 &249\\
 & &	& 	&100.0\%&	248	&245&	4.6&	245&	245 \\\midrule
\multirow{9}{*}{\rotatebox{90}{\textbf{Trade Union }}} & \multirow{4}{*}{\rotatebox{90}{\textbf{few-shot}}}&\cellcolor{lightblue}\checkmark&\cellcolor{lightblue}\checkmark&	\cellcolor{lightblue}100.0\%	&\cellcolor{lightblue}474&	\cellcolor{lightblue}473	&\cellcolor{lightblue}7.2	&\cellcolor{lightblue}473	&\cellcolor{lightblue}473\\
&&\checkmark&&	100.0\%	&505&	505&	7.7	&401&	360\\
&&&\checkmark&	95.5\%	&290&	290	&4.6&	290	&290\\
&&&&	98.5\%	&279&	279	&4.3&	279&	279\\\cmidrule{2-10}
&\multirow{4}{*}{\rotatebox{90}{\textbf{zero-shot}}}&\cellcolor{lightred}\checkmark&\cellcolor{lightred}\checkmark&	\cellcolor{lightred}100.0\%	&\cellcolor{lightred}422&	\cellcolor{lightred}420	&\cellcolor{lightred}6.4&	\cellcolor{lightred}420&	\cellcolor{lightred}420\\
&&\checkmark&&	98.5\%	&418&	418	&6.4&	324&	248\\
&&&\checkmark&	100.0\%	&299&	297	&4.5&	297&	297\\
&&&&	98.5\%	&304&	303	&4.7&	303&	303\\
\bottomrule
\end{tabular}
\end{table}

The results reveal several important trends. First, dual extractor configurations generally yield higher average terms per article and greater trilingual coverage compared to their single extractor counterparts, suggesting enhanced comprehensiveness and robustness. Second, the auto-complete strategy consistently produces higher unique term counts and broader language coverage, especially when combined with the dual extractor, indicating its utility in boosting extraction recall. Third, few-shot settings show a slight advantage in some configurations, but the gap with zero-shot is often marginal, reflecting the maturity of the model and prompt design. Lastly, the extraction results on the Trade Union Law, which contains more articles, further amplify these trends, with both dual extractor and auto-complete combinations delivering the most comprehensive and language-rich terminology lists.

Overall, the table highlights the importance of extraction strategy design. The combination of dual extractor and auto-complete strategies consistently achieves the most complete and high-coverage term extraction, which is critical for constructing multilingual legal terminology resources.

\subsection{Terminology Generation Capability}
The performance of various Large Language Models (LLMs) in extracting legal terms from the Trade Union Law, the Standardization Law and the Against Unfair Competition Law was evaluated based on several key metrics: the number of successfully processed entries (\textbf{Success}), the success rate (\textbf{Succ. Rate}), the total number of terms extracted (\textbf{Extracted}), and the number of terms after standardization (\textbf{Stand.}). The duplicate rate (\textbf{Dupl. Rate}, i.e., the ratio of duplicate terms to total extracted terms) indicates the proportion of redundant or overlapping terms among all extracted items. These models were categorized into closed-source (commercial) and open-source groups, allowing for a comparative analysis.
%各种大型语言模型（LLMs）在提取《工会法》，《标准化法》和《反不正当竞争法》中的法律术语时，基于几个关键指标进行了评估：成功处理的条目数量、成功率、提取的术语总数以及去重后的唯一术语数量。这些模型被分为闭源（专有）和开源（公开可用）两类，进行对比分析。

\begin{table}[!htbp]
\centering
\caption{Performance of dual-stream term extraction and mapping using various LLM backbones on the Standardization Law (54 articles), Trade Union Law (66 articles), and Against Unfair Competition (Against U.C., 40 articles) Law. This subset contains 160 articles in total.}\label{tab:model_comparison}
\begin{tabular}{l|l|l|rrrrr}
\toprule
\textsc{\textbf{Law}}  & & \textsc{\textbf{Model}} & \textsc{\textbf{Success}}$\uparrow$ & \textsc{\textbf{Succ. Rate}\%}$\uparrow$ & \textsc{\textbf{Extracted}}$\uparrow$ & \textsc{\textbf{Stand.}}$\uparrow$   & \textsc{\textbf{Dupl. Rate} }$\downarrow$\\
\midrule
\multirow{9}{*}{\rotatebox{90}{\textbf{Standardization}}}
 & \multirow{5}{*}{\rotatebox{90}{\textsc{Closed}}} & GPT4.1 & \textbf{54} & \textbf{100.0\%}  & 428   &  259	
 & 39.5\% \\
 & & GPT4.1-mini & 53   & 98.1\% &  427 &  286  & \textbf{33.0\%}  \\
  &  & Claude4-sonnet &   52 &  96.3\%  & 389  & 235  &   39.6\%  \\
 & & Gemini2.5-pro & 53  &  98.1\%  &  336  &   198 &  41.1\% \\	
 & & \cellcolor{lightblue}Gemini2.5-flash &  \cellcolor{lightblue}\textbf{54} &   \cellcolor{lightblue}\textbf{100.0\%}  & \cellcolor{lightblue}\textbf{466}  & \cellcolor{lightblue}\textbf{299} & \cellcolor{lightblue} 35.8\%  \\
\cmidrule{2-8}
 & \multirow{4}{*}{\rotatebox{90}{\textsc{Open}}} & Qwen3-32B &  \textbf{54}  & \textbf{100.0\% }  &  310 & 200   &  35.5\%\\
 & & Qwen3-14B & 50 & 92.6\% & 275 & 181  &  34.2\%\\
 & & Qwen3-8B & 33   & 61.1\%  & 244 & 189 & \textbf{22.5\%} \\
 & & \cellcolor{lightred}Deepseek-v3 & \cellcolor{lightred}\textbf{54} &  \cellcolor{lightred}\textbf{100.0\%}  &  \cellcolor{lightred}\textbf{356}  &  \cellcolor{lightred}\textbf{226}  & \cellcolor{lightred}36.5\%  \\
\midrule
\multirow{9}{*}{\rotatebox{90}{\textbf{Trade Union }}}
 & \multirow{6}{*}{\rotatebox{90}{\textsc{Closed}}}  
   & GPT4.1 & \textbf{66} & \textbf{100.0\%}  & 554 & 380 & 31.4\%\\
 & & \cellcolor{lightblue}GPT4.1-mini &\cellcolor{lightblue} 64 & \cellcolor{lightblue}97.0\% & \cellcolor{lightblue}\textbf{583}  & \cellcolor{lightblue}\textbf{384} &  \cellcolor{lightblue}34.1\% \\
 & & Claude4-sonnet &  62 & 93.9\%  & 453 &  286 &  36.9\% \\
 & & Gemini2.5-pro &   \textbf{66} & \textbf{100.0\%} & 461 &   275 &  40.3\%  \\
 & & Gemini2.5-flash & 65  & 98.5\%  & 527 & 362 & \textbf{31.3\%}\\
\cmidrule{2-8}
 & \multirow{4}{*}{\rotatebox{90}{\textsc{Open}}} & Qwen3-32B & 59   &  89.4\%  & 370  & 273  &  26.2\% \\
 & &  Qwen3-14B & 59   & 89.4\% & 361  & 270 & 25.2\% \\ 
 & & Qwen3-8B &  49   & 74.2\% & 331   & 262  &  \textbf{20.8\%} \\
 & & \cellcolor{lightred}{Deepseek-v3} & \cellcolor{lightred}\textbf{66} &  \cellcolor{lightred}\textbf{100.0\%}  & \cellcolor{lightred}\textbf{474}  & \cellcolor{lightred}\textbf{336}  &  \cellcolor{lightred}29.1\%  \\
\midrule 
\multirow{9}{*}{\rotatebox{90}{\textbf{Against U. C. }}}
 & \multirow{5}{*}{\rotatebox{90}{\textsc{Closed}}} & GPT4.1  & \textbf{40}& \textbf{100.0\%} &	329	& 223  & 32.2\% \\
 & & GPT4.1-mini &39  &	97.5\% &	342	& 	\textbf{238}	&\textbf{30.4\%} \\
 & & Claude4-sonnet & 	39 & 97.5\% & 274	&176  & 35.8\% \\
 & & Gemini2.5-pro & \textbf{40} &	\textbf{100.0\%}	&291&	177  & 39.2\% \\ 
 & & \cellcolor{lightblue}{Gemini2.5-flash}& \cellcolor{lightblue}39 &	\cellcolor{lightblue}97.5\% & \cellcolor{lightblue}\textbf{362} & \cellcolor{lightblue}237 & \cellcolor{lightblue} 34.5\%\\
\cmidrule{2-8}
 & \multirow{4}{*}{\rotatebox{90}{\textsc{Open}}} & Qwen3-32B &	37   &  92.5\% &  205 & 133  &  35.1\% \\
 & & Qwen3-14B & 36 & 90.0\%  &200 & 133 & 33.5\% \\
 & & Qwen3-8B  &  27  & 67.5\% &  193 & 139 &  \textbf{28.0\%} \\
 & & \cellcolor{lightred}Deepseek-v3 & \cellcolor{lightred}\textbf{40}  & \cellcolor{lightred}\textbf{100.0\%} & \cellcolor{lightred}\textbf{267} & \cellcolor{lightred}\textbf{177} &  \cellcolor{lightred}33.7\% \\
\midrule 
\multirow{9}{*}{\rotatebox{90}{\textbf{Total }}}
 & \multirow{5}{*}{\rotatebox{90}{\textsc{Closed}}} & GPT4.1  & \textbf{160}& \textbf{100.0\%} &	1,311	& 862  & 34.2\% \\
 & & GPT4.1-mini & 156	 &	97.5\% &	1,307	& 	\textbf{908}	&\textbf{30.5\%} \\
 & & Claude4-sonnet & 153	 &  95.6\% & 1,116  & 697 &  37.5\% \\		
 & & Gemini2.5-pro & 159 &	99.3\% 	&1,088&	  650 & 40.3\% \\ 
 & & \cellcolor{lightblue}{Gemini2.5-flash}& \cellcolor{lightblue}158  &	\cellcolor{lightblue}\textbf{98.8\%} & \cellcolor{lightblue}\textbf{1,355} & \cellcolor{lightblue}898 & \cellcolor{lightblue}33.7\%\\
\cmidrule{2-8}
 & \multirow{4}{*}{\rotatebox{90}{\textsc{Open}}} & Qwen3-32B &	150  & 	93.8\% &  885 & 606 & 31.5\% \\	
 & & Qwen3-14B & 145 & 90.6\%  &836 & 584 & 30.1\% \\
 & & Qwen3-8B  &  109  & 68.1\% &  768 & 590 &  \textbf{23.2\%} \\
 & & \cellcolor{lightred}Deepseek-v3 & \cellcolor{lightred}\textbf{160}  & \cellcolor{lightred}\textbf{100.0\%} & \cellcolor{lightred}\textbf{1,097} & \cellcolor{lightred}\textbf{739} &  \cellcolor{lightred}32.6\% \\ 
\bottomrule 
\end{tabular}
\end{table}

Table~\ref{tab:model_comparison} presents a cross-law and cross-model comparison of term extraction performance using the few-shot dual extractor with auto completion strategy. The table summarizes results for three representative legal codes, the Standardization Law (2017), the Trade Union Law (2001), and the Against Unfair Competition Law (2019), across a range of leading large language models (LLMs), including Qwen3 (in 8b, 14b, 32b sizes), GPT4.1 and GPT4.1-mini, Gemini2.5-pro and Gemini2.5-flash, Deepseek-v3, and Claude-4-sonnet.
%\textbf{表~\ref{tab:model_comparison}} 展示了在 few-shot dual extractor + auto completion 策略下，不同大模型在三部法律（《标准化法(2017)》、《工会法(2001)》和《反不正当竞争法(2019)》）上的术语提取效果对比。涵盖的主流大模型包括 Qwen3（8b、14b、32b）、GPT4.1、GPT4.1-mini、Gemini2.5-pro、Gemini2.5-flash、Deepseek-v3 和 Claude-4-sonnet。

Several trends are evident from the results. First, the larger models generally achieve higher success rates in article processing, with models such as GPT4.1, Gemini2.5-flash, and Deepseek-v3 reaching nearly 100\% across all laws. However, the number of unique terms extracted varies significantly between models. For example, GPT4.1 and Gemini2.5-flash consistently yield higher total and unique term counts, and also display greater trilingual coverage, as indicated by the number of articles with Chinese, Japanese, and English equivalents.
%结果显示，较大规模模型通常在条文处理成功率上表现更好，如 GPT4.1、Gemini2.5-flash 和 Deepseek-v3 等模型在各部法律上的处理成功率几乎均达到了 100%。然而，不同模型在唯一术语数以及去重率（即 unique/total extracted）的表现差异较大。例如，GPT4.1 和 Gemini2.5-flash 在总术语数与唯一术语数上始终位居前列，且三语覆盖度（即含中文、日文、英文术语条目数量）表现也十分突出。
Second, while smaller models (e.g., Qwen3-8b) sometimes achieve competitive performance in terms of average terms per article, they tend to lag in both extraction coverage and duplicate rate compared to their larger counterparts. Notably, some models, such as Gemini2.5-pro and Claude-4-sonnet, achieve the highest duplicate rates (exceeding 35--40\%), indicating more redundant or overlapping terms.
%此外，部分小模型（如 Qwen3-8b）在每条平均术语数方面有一定竞争力，但在术语覆盖率和去重率上仍落后于大模型。值得注意的是，部分模型如 Gemini2.5-pro 和 Claude-4-sonnet 去重率最高，均超过 35--40%，显示出较强的冗余术语过滤和去重能力。
Third, the variability in performance across different laws highlights the importance of legal domain and text structure in extraction outcomes. The Trade Union Law, with more articles, allows for greater overall term coverage and more reliable comparison of model behaviors. In contrast, performance fluctuations are more pronounced in shorter legal texts.
%不同法律的实验结果也反映出法律文本结构对抽取效果的影响。例如，条文数更多的《工会法》能够更充分地展现各模型的提取能力，而条文较少的《反不正当竞争法》模型间的波动则更明显。

Overall, the table demonstrates that recent LLMs, when paired with a robust extraction pipeline, can deliver highly comprehensive and multilingual terminology resources. However, the duplicate rate and the balance between extraction breadth and precision remain key challenges. Future work should further explore hybrid approaches to maximize both term coverage and uniqueness. Among all evaluated models, \textbf{GPT4.1}, \textbf{GPT4.1-mini}, \textbf{Gemini2.5-flash}, and \textbf{Deepseek-v3} achieve the good success rates, extracting over 1,000 terms each with strong trilingual coverage. Additionally, given the cost and API prices, \textbf{GPT4.1-mini}, \textbf{Gemini2.5-flash}, and \textbf{Deepseek-v3} are recommended for high-quality, comprehensive, and precise multilingual legal terminology extraction.
%总体来看，结合强大的大语言模型和鲁棒抽取策略，能够高效获得高质量、三语齐全的法律术语资源。但如何进一步提升去重率，实现术语覆盖和唯一性之间的最优平衡，仍是后续研究的重要方向。
\subsection{LLM-based Terminology Quality Assessment}
Due to the open-ended nature of the task, we cannot predefine how many target or pivot language terms correspond to each source language term, making it difficult to construct a fixed ``gold standard'' to measure the correctness and completeness of each output. In this context, traditional methods of calculating precision, recall, and F1 score based on a fixed answer set are no longer applicable. Applying these metrics would not only fail to accurately reflect the system's performance in a real-world open environment, but it could also obscure the unique complexities and contextual sensitivity of multilingual legal terminology.

To address this, we adopted a more adaptive evaluation approach: a multi-dimensional analysis framework combining large language models with domain experts to perform both qualitative and quantitative evaluations of the output. This approach takes into account multiple dimensions, including linguistic, contextual, and legal expertise, aligning more closely with the flexibility and depth of judgment required for open-ended terminology extraction tasks. We are not dismissing the value of traditional metrics; rather, based on the nature of the task and the data characteristics, we have chosen a more interpretable and practical approach to ensure the rigor and feasibility of the evaluation.

\subsubsection{Evaluation Metrics}
This study has designed a comprehensive evaluation framework for multilingual legal terminology mappings, which aims to conduct a thorough and systematic evaluation through five core dimensions. As shown in  Table~\ref{tab:termbase_metrics}, this framework combines the intelligent judgment capabilities of large language models and objective quantitative metrics based on statistical analysis, ensuring the accuracy, reliability, and operability of the evaluation results. The evaluation of terminology quality in the context of multilingual legal texts involves a multi-perspective approach, addressing several key dimensions such as coverage, consistency, completeness, professionalism, and translation quality. Below, we outline the methodology adopted for assessing these components.

In the evaluation of terminology lists, we conduct sample-based quality assessment on a termbase $T = {t_1, t_2, ..., t_N}$ of total size $N$.
The sampling strategy is defined as follows: when $N > 100$, we adopt a sequential sampling approach and randomly select $n = 100$ terms as the evaluation sample $S = {t_1, t_2, ..., t_{100}}$; when $N \leq 100$, we use full-sample evaluation, i.e., $S = T$. This method ensures both the representativeness of the evaluation and efficient control over computational cost and LLM input length constraints. 

Finally, the overall quality score \( Q \) for the terminology set is computed by aggregating the individual scores for each aspect as $\sum_{i=1}^{5} w_i \cdot M_i$
% :
% \begin{equation}
% Q = \sum_{i=1}^{5} w_i \cdot M_i \quad 
% \end{equation}
% Where \({ w_1, w_2, w_3, w_4, w_5 }\) are 
with weights assigned to each dimension based on its importance in the context of the specific legal system being evaluated and $M_i$ belongs to the set of \{\text{Coverage}, \text{Consistency}, \text{Completeness}, \text{Professionalism}, \text{Translation Quality}\}.

\begin{sidewaystable}
\footnotesize
\caption{\small TermBase systematic evaluation metrics (LLM-centric prompt-based framework)}
\label{tab:termbase_metrics}
\begin{tabular}{p{0.6cm}|p{3cm}|p{7.2cm}|p{5.2cm}|p{1cm}}
\toprule
\textbf{Dim.}  & \textbf{Sub-aspect.} & \textbf{Detailed Information (LLM Task)} & \textbf{Design Rationale} & \textbf{Weight} \\
\midrule

\multirow{9}{*}{\rotatebox{90}{\textbf{Coverage}}} 
& Semantic Coverage & List all unique legal concepts in this termbase. Identify and count redundant or duplicate entries. & Prevents terminology redundancy and ensures conceptual diversity within the legal domain & \multirow{3}{*}{\quad \textbf{25\%}} \\
\cmidrule{2-4}
& Legal Domain Coverage & Classify terms into legal sub-domains. Report domain coverage and the ratio of general vs. specialized legal vocabulary. & Ensures comprehensive domain coverage and inclusion of professional legal categories & \\
\cmidrule{2-4}
& Term Diversity & Evaluate the lexical diversity and variety of expressions among all terms. Are the terms evenly distributed, or do they show repetition? & Measures richness and prevents monotony in legal vocabulary & \\
\midrule

\multirow{11}{*}{\rotatebox{90}{\textbf{Consistency}}}
& Translation Consistency & For each source term, check whether multiple translations exist in the same language. Report excessive variation and assess if variants are justified by polysemy. & Balances translation flexibility and consistency; prevents excessive chaos & \multirow{4}{*}{\quad \textbf{25\%}} \\
\cmidrule{2-4}
& Terminology System & Evaluate if the terminology set forms a logical hierarchy with clear naming conventions and professional classifications. & Ensures systematic organization of legal knowledge & \\
\cmidrule{2-4}
& Format Standardization & Assess whether all required fields are consistently completed and if the field naming follows a uniform format. & Improves usability and data quality & \\
\cmidrule{2-4}
& Semantic Consistency & For each term and its translations, check if the meanings are aligned and accurate across languages, and whether context is consistently maintained. & Ensures cross-lingual semantic accuracy and coherence & \\
\midrule

\multirow{8}{*}{\rotatebox{90}{\textbf{Completeness}}}
& Information Richness & Check if each term entry contains all mandatory and value-added information fields (definition, context, source, explanation, etc.). Score the richness. & Encourages comprehensive, informative entries & \multirow{3}{*}{\quad \textbf{20\%}} \\
\cmidrule{2-4}
& Translation Completeness & Verify that every entry has translations in all required languages. Flag missing or incomplete translations. & Ensures completeness of multilingual coverage & \\
\cmidrule{2-4}
& Contextual Completeness & Assess if each term provides sufficient context, explanation, and source information for effective understanding and use. & Ensures terms are understandable and practically useful & \\
\midrule

\multirow{10}{*}{\rotatebox{90}{\textbf{Professionalism}}}
& Linguistic Quality & Evaluate whether terms are of appropriate length, use professional legal vocabulary, and are well-formed. & Maintains linguistic and academic quality & \multirow{4}{*}{\quad \textbf{15\%}} \\
\cmidrule{2-4}
& Professional Standard & Check if the term and translation conform to legal industry standards and authoritative references. & Ensures professional credibility & \\
\cmidrule{2-4}
& Usability & Assess if terms and explanations are practical, easy to understand, and suitable for search and real-world application. & Promotes practical applicability & \\
\cmidrule{2-4}
& Legal Terminology Accuracy & Verify that each entry accurately represents a legal concept and uses correct legal language. & Ensures legal precision & \\
\midrule

\multirow{8}{*}{\rotatebox{90}{\quad \textbf{Trans. Quality}}}
& Chinese-Japanese Quality & Evaluate the quality of Chinese-Japanese translations in terms of accuracy, professionalism, naturalness, and consistency. & Ensures professional translation between Chinese and Japanese legal terms & \multirow{3}{*}{\quad \textbf{15\%}} \\
\cmidrule{2-4}
& Chinese-English Quality & Evaluate the quality of Chinese-English translations in terms of accuracy, professionalism, naturalness, and consistency. & Ensures professional translation between Chinese and English legal terms & \\
\cmidrule{2-4}
& Translation Naturalness & Assess whether translations are idiomatic and natural in the target language, beyond mere literal correctness. & Ensures translation fluency and usability & \\
\bottomrule
\end{tabular}
\end{sidewaystable}

\subsubsection{LLM-based Evaluation}

Besides, the multi-agent terminology evaluation system leverages advanced large language models (LLMs) to simulate the judgment of expert linguists and legal professionals across multiple quality dimensions. For each trilingual legal terminology database, five specialized evaluation agents—each powered by an LLM—independently assess a distinct aspect of quality: coverage, consistency, completeness, professionalism, and translation accuracy. Each agent automatically reviews sampled entries and relevant metadata, generating an objective score (ranging from 0 to 100) for its respective dimension based on predefined evaluation criteria and prompt instructions.

To ensure efficiency and scalability, all five LLM agents operate in parallel, rapidly processing and scoring thousands of legal terms in a fraction of the time required for manual review. The individual scores from each agent are then aggregated using a weighted formula to produce a comprehensive overall grade (e.g., A+, A, B, etc.) for each terminology database. This approach combines the nuanced judgment capabilities of state-of-the-art LLMs with systematic, reproducible evaluation protocols, enabling rapid, expert-level quality assessment of large multilingual legal terminology resources. 

Table~\ref{tab:llmscore} presents a comprehensive evaluation of legal terminology standardization quality across three representative legal datasets, utilizing multiple leading large language models (LLMs) as independent evaluators. Each row represents the results for a specific model, while the columns summarize scores for five core metrics: Coverage (\textbf{Cov.}), Consistency (\textbf{Cons.}), Completeness (\textbf{Comp.}), Quality (\textbf{Prof.}), and Translation Quality (\textbf{Trans.}), and overall score and grade.

\begin{table}[!htbp]
\centering
\caption{LLM-self evaluation for the quality of the terminology extraction, mapping, and standardization ({\color{lightblue}{Gemini2.5-pro}} and {\color{lightred}{GPT4.1}} have the best performance on this task).}
\label{tab:llmscore}
\begin{tabular}{l|l|rrrrr|cl}
\toprule
 &\textbf{Model} &  \textbf{Cov.} & \textbf{Cons.} & \textbf{Comp.} & \textbf{Prof.} & \textbf{Trans.}  & \textbf{Score}$\uparrow$ & \textbf{Grade}$\uparrow$  \\
\toprule 
\multirow{9}{*}{\rotatebox{90}{\textbf{DeepSeek-v3}}}
&Gemini2.5-flash   & 85 & 87 & 99 & 97 & 88 & 91.85 & A \\
&\cellcolor{lightblue}Gemini2.5-pro     & \cellcolor{lightblue}85 & \cellcolor{lightblue}87 & \cellcolor{lightblue}100 & \cellcolor{lightblue}91 & \cellcolor{lightblue}88 & \cellcolor{lightblue}91.25 & \cellcolor{lightblue}A \\
&Claude-4-sonnet   & 85 & 87 & 100 & 89 & 88 & 90.95 & A \\
&GPT4.1-mini       & 85 & 87 & 100 & 87 & 88 & 90.65 & A \\
&\cellcolor{lightred}GPT4.1            & \cellcolor{lightred}85 & \cellcolor{lightred}89 & \cellcolor{lightred}97  & \cellcolor{lightred}89 & \cellcolor{lightred}88 & \cellcolor{lightred}90.45 & \cellcolor{lightred}A \\
&DeepSeek-v3       & 85 & 87 & 100 & 87 & 82 & 89.75 & A- \\
&Qwen3-32b         & 85 & 85 & 95  & 89 & 88 & 89.05 & A- \\
&Qwen3-14b         & 85 & 87 & 77  & 89 & 88 & 84.05 & B+ \\
&Qwen3-8b          & 85 & 85 & 75  & 89 & 82 & 82.15 & B+ \\
\midrule
\multirow{9}{*}{\rotatebox{90}{\textbf{GPT4.1}}}
&\cellcolor{lightblue}Gemini2.5-pro   & \cellcolor{lightblue}92 & \cellcolor{lightblue}90 & \cellcolor{lightblue}98  & \cellcolor{lightblue}93 & \cellcolor{lightblue}96 & \cellcolor{lightblue}94.15 & \cellcolor{lightblue}A \\
&\cellcolor{lightred}GPT4.1          & \cellcolor{lightred}86 & \cellcolor{lightred}93 & \cellcolor{lightred}97  & \cellcolor{lightred}92 & \cellcolor{lightred}96 & \cellcolor{lightred}93.10 & \cellcolor{lightred}A \\
&GPT4.1-mini     & 91 & 84 & 98  & 84 & 95 & 91.25 & A \\
&Gemini2.5-flash & 88 & 93 & 88  & 94 & 97 & 91.25 & A \\
&Claude-4-sonnet & 84 & 84 & 100 & 93 & 91 & 91.20 & A \\
&DeepSeek-v3     & 92 & 64 & 100 & 92 & 91 & 88.65 & A- \\
&Qwen3-32b       & 91 & 64 & 97  & 60 & 92 & 82.90 & B+ \\
&Qwen3-8b        & 84 & 64 & 80  & 81 & 93 & 79.70 & B \\
&Qwen3-14b       & 84 & 60 & 60  & 90 & 93 & 74.25 & B- \\ \midrule
\multirow{9}{*}{\rotatebox{90}{\textbf{Gemini2.5-pro}}}
&\cellcolor{lightblue}Gemini2.5-pro   & \cellcolor{lightblue}55 & \cellcolor{lightblue}70 & \cellcolor{lightblue}100 & \cellcolor{lightblue}83&\cellcolor{lightblue}82& \cellcolor{lightblue}79.75 & \cellcolor{lightblue}B \\
&\cellcolor{lightred}GPT4.1          & \cellcolor{lightred}62 & \cellcolor{lightred}63 & \cellcolor{lightred}100 & \cellcolor{lightred}76&\cellcolor{lightred}60 & \cellcolor{lightred}75.40 & \cellcolor{lightred}B \\
&Gemini2.5-flash & 68 & 66 & 90  & 50& 65 & 71.05 & B- \\
&Claude-4-sonnet & 67 & 58 & 100 & 60&40 & 70.00 & B- \\
&GPT4.1-mini     & 62 & 47 & 100 & 50&60 & 68.30 & C+ \\
&Qwen3-32b       & 60 & 60 & 100 & 50&35 & 66.75 & C+ \\
&DeepSeek-v3     & 60 & 34 & 100 & 35&25 & 63.05 & C \\
&Qwen3-8b        & 77 & 20 & 80  & 31& 20& 57.05 & C- \\
&Qwen3-14b       & 68 & 22 & 90  & 20&25& 51.75 & C- \\
\bottomrule
\end{tabular}
\end{table} 
This study conducted a systematic comparison of nine mainstream large language models (LLMs) on multilingual legal terminology extraction tasks, utilizing both self-evaluation and cross-evaluation frameworks. The experimental results reveal notable differences in scoring patterns across the three main evaluators: DeepSeek-v3, GPT4.1, and Gemini2.5-pro. Both DeepSeek-v3 and GPT4.1 exhibit a generally lenient scoring trend, awarding most leading models (such as Gemini2.5, Claude-4-sonnet, and GPT4.1-mini) with A or A- grades. In contrast, Gemini2.5-pro adopts a significantly stricter evaluation standard, with all models (including itself) receiving only B or lower grades. Across all evaluation systems, top-performing models such as \textbf{Gemini2.5-pro} and \textbf{GPT4.1} consistently achieve high scores for coverage, consistency, completeness, and domain-specificity, demonstrating robust capacity in extracting and aligning legal terms across languages. The completeness metric is especially high across the board, suggesting strong performance in term coverage and contextual fidelity. However, there is greater variability in consistency and domain-specificity, where models such as Qwen3-8b and Qwen3-14b exhibit notable weaknesses.

A further observation is the clear bias present in self-evaluations: both DeepSeek-v3 and GPT4.1 tend to rate themselves and similar models more generously, whereas Gemini2.5-pro's self-assessment is markedly conservative. This disparity underscores the influence of differing evaluation philosophies and quality standards among developers. It also highlights the limitations of relying solely on a single model's self-assessment, reinforcing the necessity for cross-evaluation and expert human review to achieve more objective and reliable benchmarking.

Overall, our multi-model, multi-dimensional evaluation framework provides valuable insights into the strengths and weaknesses of each LLM in legal terminology extraction. It also offers a solid foundation for future research on automated legal termbase construction. Further work should emphasize more granular analysis of sub-metrics and deeper integration of human expert assessment to promote the development of high-quality, multilingual legal resources.

\subsection{Human Terminology Quality Assessment}

For human evaluation, five expert reviewers with backgrounds in legal translation and multilingual terminology independently assessed nine multilingual language models. For each model, 100 trilingual legal terms were randomly sampled from the generated terminology tables. The evaluation followed five criteria: coverage, consistency, completeness, professionalism, and translation quality. 

\begin{figure}[!htbp]
\centering
\includegraphics[width=\textwidth]{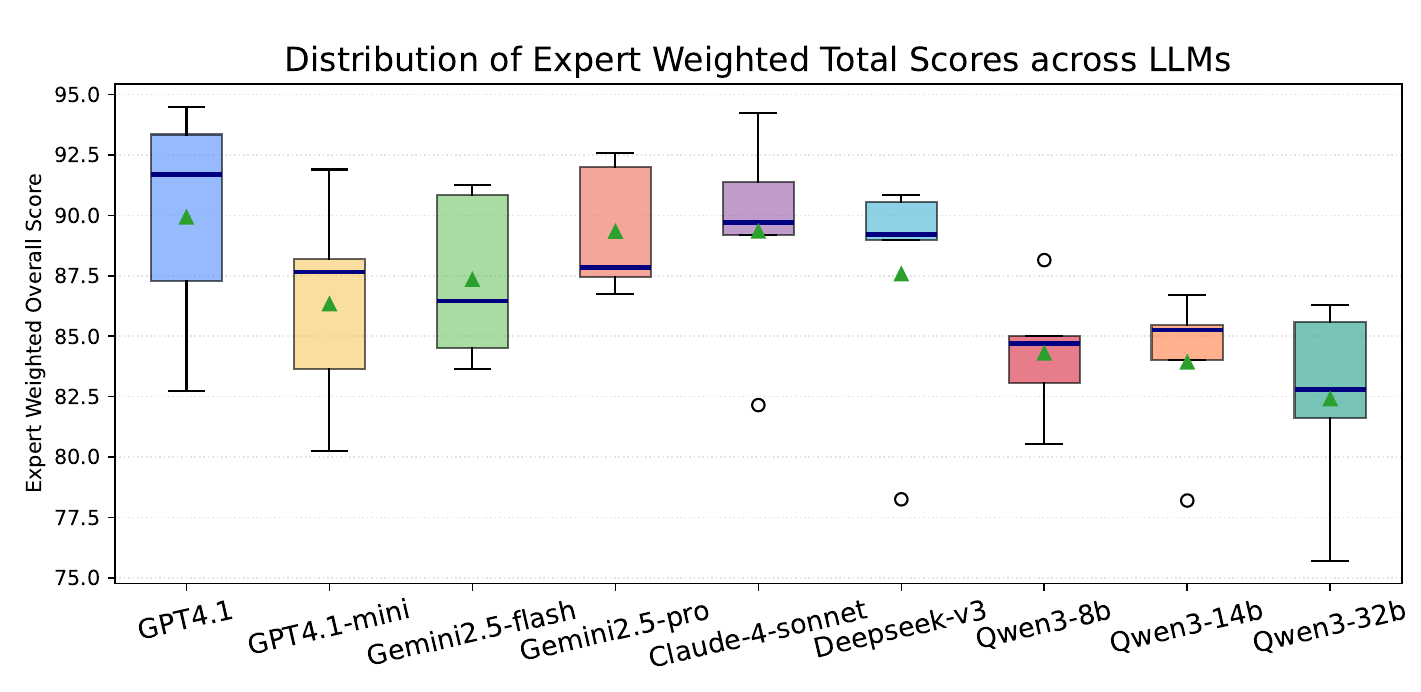}
\caption{Human evaluation of expert weighted overall scores.}\label{boxplot}
\end{figure}

\begin{figure}[!htbp]
\centering
\includegraphics[width=\textwidth]{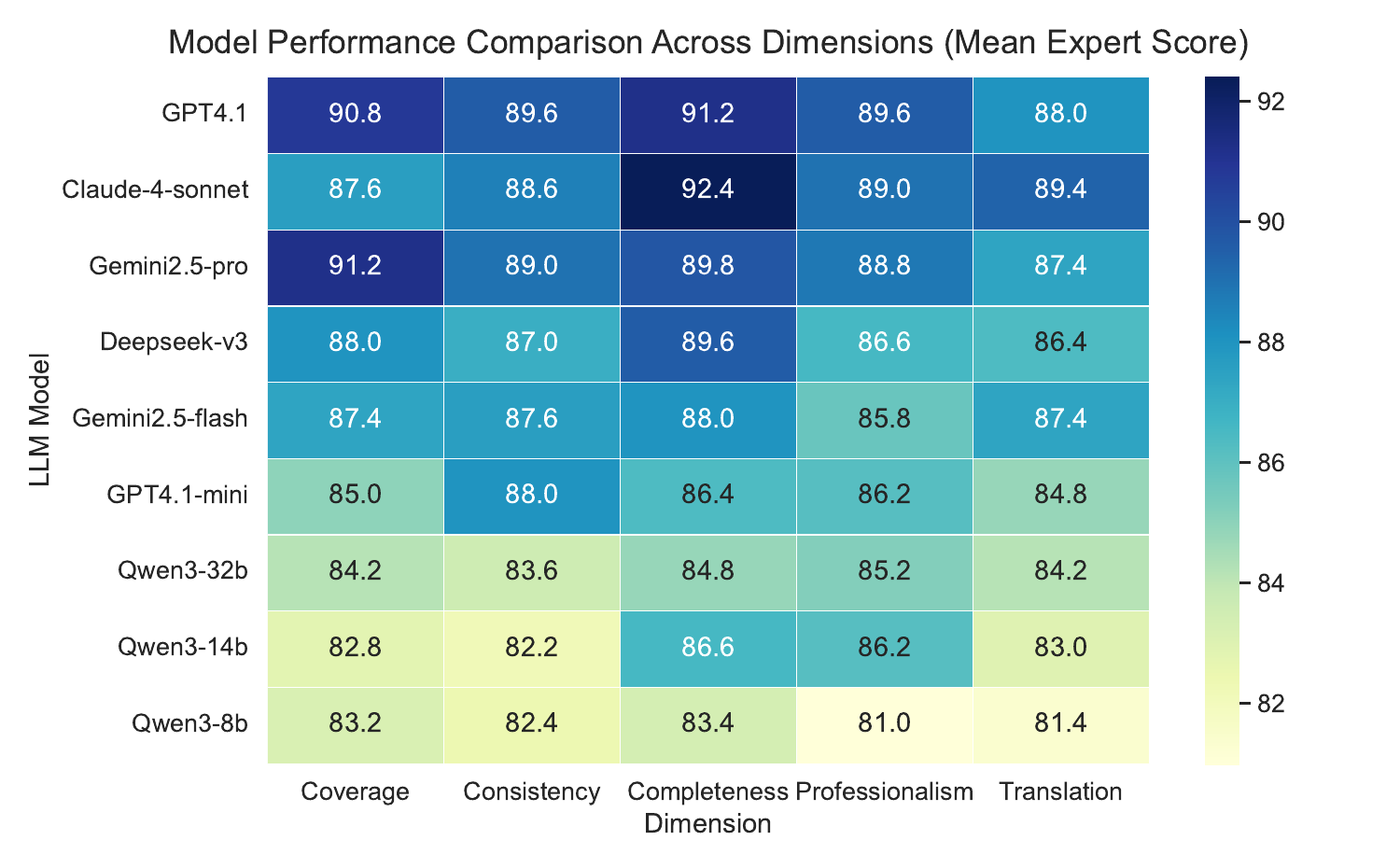}
\caption{Human evaluation for the quality of multilingual legal terminology mapping across different evaluation dimensions.}\label{hotmap}
\end{figure}

Figure~\ref{boxplot} illustrates the distribution of scores assigned by individual human experts for each model. The Figure~\ref{hotmap} summarizes the mean expert scores across different evaluation dimensions, providing a clear comparison of model performance by criterion. 

Claude-4-sonnet and GPT4.1 demonstrated the best overall performance, with consistent strengths in completeness, context accuracy, and natural legal phrasing. Gemini2.5-pro and DeepSeek-v3 followed closely, delivering reliable results across most criteria, then Gemini2.5-flash and GPT4.1-mini. All smaller Qwen3 variants lagged behind in completeness and translation quality due to more frequent omissions and less authoritative definitions. Top-rated models were distinguished by comprehensive definitions, robust trilingual alignment, and idiomatic translations. Lower-rated outputs, notably from 8b, were affected by missing context, literal translation, or informal wording.
Overall, Claude-4-sonnet, GPT4.1, and Gemini2.5-pro are recommended for multilingual legal terminology extraction where completeness and linguistic quality are required.

\subsection{Summary of Extraction Results}
The extraction and processing of the trilingual parallel termbase have yielded remarkable results. Through an automated workflow, 18,845 high-quality Chinese-Japanese-English legal term entries were generated from 41,423 original entries, achieving comprehensive coverage of core legal concepts. Leveraging large language models, 22,578 synonymous or near-synonymous variants were intelligently merged, resulting in an overall merging efficiency of 86.1\% and a standardization rate of 98.4\%. All entries retained complete trilingual information, ensuring both data integrity and independence, with 18,281 unique Chinese terms representing a 97.0\% independence rate. Meanwhile, all variants with semantic differences were effectively identified and preserved, avoiding any loss of meaning or ambiguity. Automated deduplication further enhanced the usability and searchability of the termbase, achieving a total data reduction rate of 54.5\%. The entire pipeline from term extraction and variant identification to standardization, language alignment, and data evaluation was fully automated, significantly reducing manual workload. Multidimensional quality assessments demonstrated outstanding performance in terms of completeness, accuracy, consistency, usability, and intelligence. This work provides a robust data foundation for multilingual legal text processing and intelligent translation, while laying the groundwork for ongoing dynamic maintenance and further enhancement of the termbase. The samples of our constructed termbase can be found in Appendix~\ref{secC}.

\subsection{Case Study}
Through a close examination of the aligned entries, we identified four primary categories of challenges commonly encountered in multilingual legal terminology resources: variants, redundancy, context mismatch or over-extraction, and hallucinations. It is important to note that these issues can largely be mitigated by following the optimized prompts and extraction guidelines we propose for large language models. While our approach significantly reduces the occurrence of such problems, occasional errors remain inevitable due to the inherent limitations of current AI models. Therefore, we strongly recommend adopting our standardized prompts and best practices to maximize quality and reliability in multilingual legal terminology extraction.

Therefore, further terminology standardization  or quality assurance by legal experts is necessary to address these issues and enhance the usability of the extracted data, which involves unifying translation variants, assigning a unique identifier to each legal concept, clustering term variants by semantic equivalence, and designating a canonical form for each term group. Moreover, context information can be categorized or labeled to support more precise mapping between terms and legal provisions. These steps are essential to improve the quality, consistency, and interoperability of the multilingual legal terminology resource, and will form the focus of our subsequent standardization work.

\subsubsection{Variants}
These inconsistencies in wording, capitalization, and the granularity of translation (ranging from full legal titles to action phrases) complicate downstream tasks such as automated alignment and knowledge base construction.

\begin{table*}[!htbp]
\centering
\caption{Examples of multilingual legal term variants in the Against Unfair Competition (Against U.C.) Law.}
\label{tab:legal_terms_redundancy_examples}
\resizebox{\textwidth}{!}{\small
\begin{tabular}{p{3cm} p{2.5cm} p{3cm} p{5cm}}
\toprule
\textbf{Chinese} & \textbf{Japanese} & \textbf{English} & \textbf{Context (en)} \\
\toprule
不正当竞争行为 & 不正競争行為 & acts of unfair competition &  preventing acts of unfair competition \\\midrule
不正当竞争行为 & 不正競争行為 & Acts of Unfair Competition &  Chapter II Acts of Unfair Competition \\\midrule
不正当竞争行为 & 不正競争行為 & act of unfair competition & For the purposes of this Law, 'an act of unfair competition' \\ \midrule
不正当竞争行为 & 不正競争行為 & unfair competition acts &  engage in public supervision over unfair competition acts \\\midrule
不正当竞争行为 & 不正競争が疑われる行為 & acts of unfair competition &  Investigation into Suspected Acts of Unfair Competition \\\midrule
不正当竞争行为 & 不正競争が疑われる行為 & act of unfair competition &  report a suspected act of unfair competition \\
\bottomrule
\end{tabular}}
\end{table*}

Table~\ref{tab:legal_terms_redundancy_examples} presents a variety of translations and term variants for the same legal concept. Multiple translation variants and repetitive forms for the same legal concept, such as ``acts of unfair competition,'' ``act of unfair competition,'' and ``unfair competition acts.'' Additionally, inconsistencies in capitalization, the presence of both full legal titles and granular action terms, and the use of explanatory rather than strictly parallel translations complicate downstream processing, such as machine translation and knowledge base construction.
\subsubsection{Redundancy}
\begin{table*}[!htbp]
\centering
\caption{Examples of multilingual legal term redundancy in the Standardization Law and Trade Union Law.}
\label{tab:legal_terms_examples}
\resizebox{\textwidth}{!}{\small
\begin{tabular}{p{3cm} p{3cm} p{6cm} p{2.5cm}}
\toprule
\textbf{Chinese} & \textbf{Japanese} & \textbf{English} & \textbf{Source Law}  \\
\toprule
企业 & 企業 & enterprises &  Standardization \\\midrule
企业、事业单位&	企業、事業体	&enterprises and institutions&  Trade Union  \\\midrule
企业、事业单位 & 企$\cdot$事業体 & enterprises and public institutions & Trade Union  \\ \midrule
企业、事业单位、机关 & 企$\cdot$事業体、機関 & enterprises, public institutions, and government agencies &    Trade Union \\\midrule
企业、社会团体和教育、科研机构等 & 企業、社会団体、教育$\cdot$科学研究機関等 & enterprises, social organizations, educational institutions, research institutes and other organizations & Standardization\\
\bottomrule
\end{tabular}}
\end{table*}
Table~\ref{tab:legal_terms_examples} demonstrates a typical form of necessary redundancy in multilingual legal terminology extraction. While the entries (such as ``enterprises,'' ``enterprises and public institutions,'' and ``enterprises, public institutions, and government agencies'') may appear repetitive, this incremental listing serves important legal functions. Such redundancy reflects the drafting practices in legal documents, where enumerating related entities with varying levels of specificity ensures clarity, inclusiveness, and legal precision. In cross-lingual and cross-jurisdictional contexts, these distinctions are often preserved or further elaborated in translation to capture all relevant legal nuances.

However, this necessary redundancy presents challenges for automated terminology extraction and database management. Treating every variant as an independent term can fragment terminology resources and complicate downstream processing. Therefore, it is essential to balance legal accuracy with computational efficiency: standardizing terms by clustering semantically equivalent variants under unified concepts, while retaining the nuanced distinctions required for legal interpretation and practical use.

\subsubsection{Context Mismatch and Over-Extraction}

\begin{table*}[!htbp]
\centering
\caption{Examples of multilingual legal term mismatch and over-extraction.}
\label{tab:legal_terms_examples_overextraction}
\resizebox{\textwidth}{!}{\small
\begin{tabular}{p{2cm} p{2cm} p{4cm} p{4cm}  p{1cm}}
\toprule
\textbf{Chinese} & \textbf{Japanese} & \textbf{English} & \textbf{Context(en)} & \textbf{LLM} \\
\toprule
可替换性 & 代替可能性 & interoperability {\color{red}{of products}} &  enhance the ... interoperability of products & GPT4.1 \\ \midrule
处理结果	& 処理結果	&result of its {\color{red}{investigation and}} handling & the relevant regulatory department shall notify the informant of the result of its investigation and handling  & Gemini2.5-pro\\
\bottomrule
\end{tabular}}
\end{table*}
Table~\ref{tab:legal_terms_examples_overextraction} shows the cases of multilingual term mismatch and over-extraction. This phenomenon is primarily attributable to structural differences between Chinese and English legal texts. When extracting terms from Chinese provisions, large language models (LLMs) sometimes misinterpret sentence boundaries or syntactic roles, resulting in the extraction of not only the intended legal terms but also intervening phrases or contextual fragments. For example, components embedded in the middle of a sentence, such as procedural details or subordinate clauses—may be mistakenly identified as standalone terms. This over-extraction is further amplified by the relative lack of explicit word boundaries in Chinese and the complex syntactic segmentation required for accurate English mapping. Consequently, the resulting English terminology set may contain redundant or incomplete phrases, which can hinder the standardization and interoperability of multilingual legal terminology databases. Addressing this issue requires more refined boundary detection algorithms and, in many cases, expert post-processing to filter out non-essential fragments.

\subsubsection{Hallucinations}

\begin{table*}[!htbp]
\centering
\caption{Examples of hallucinations in multilingual legal term extraction.}
\label{tab:legal_terms_examples_hallucination}
\resizebox{\textwidth}{!}{\small
\begin{tabular}{p{1cm} p{1.5cm} p{2cm} p{2.5cm} p{2.5cm}p{3cm} p{1.5cm}}
\toprule
\textbf{Chinese} & \textbf{Japanese} & \textbf{English} & \textbf{Context(zh)} & \textbf{Context(ja)} & \textbf{Context(en)} & \textbf{Law/LLM} \\
\toprule
{\color{red}{商业宣传幇助}} & {\color{red}{商業宣伝の幇助}} & {\color{red}
{helping commercial publicity}}  &  经营者不得通过组织虚假交易等方式，帮助其他经营者进行虚假或者引人误解的商业宣传  &その他の事業者が虚偽の、又は関連公衆に誤解を生じさせる商業宣伝を行うことを幇助してはならない	  & shall not help another business entity engage in any false or misleading commercial publicity by organizing a false transaction or by any other means & Against U.C \newline /GPT4.1-mini \\ \midrule
实名举报人 & {\color{red}実名通報者} &	{\color{red}real-name reporter} & 	对实名举报人或者投诉人& 	通報者又は苦情申立人の実名での通報、苦情申立てについて	& reports or complaints from people using their real names &  Standard-\newline ization \newline /gemini2.5-flash\\
\bottomrule
\end{tabular}}
\end{table*}

In Table~\ref{tab:legal_terms_examples_hallucination}, the terms ``商业宣传幇助'', ``商業宣伝の幇助'', ``helping commercial publicity'' in the first example and ``実名通報者'', ``real-name reporter'', do not exist in the real context. 
LLM-based extraction occasionally generates hallucinated terms—i.e., terms or translations not present in the original legal text or not supported by legal context. These include inappropriate literal translations, over-generalizations, or the fabrication of legal terms that lack statutory basis, which can undermine the reliability of the terminology resource.

To address these challenges, further standardization and quality control measures are necessary, including the unification of term variants, assignment of unique concept identifiers, clustering of semantically equivalent terms, designation of canonical forms, and context-aware labeling. These steps are essential to ensure the reliability, consistency, and practical utility of the multilingual legal terminology database.
\begin{table}[!htbp]
\centering
\caption{Statistics of the hallucination rate.}
\begin{tabular}{l|rr|rr|rr}
\toprule
\textbf{Model} & \multicolumn{2}{c|}{\textbf{Standardization}} & \multicolumn{2}{c|}{\textbf{Trade Union}} & \multicolumn{2}{c}{\textbf{Against U. C.}}  \\ \midrule
 & Spurious/Total & Ratio  & Spurious/Total & Ratio  & Spurious/Total & Ratio \\
\midrule
GPT4.1 & 11/428 &2.6\%  & 39/554 & 7.0\% &  6/329& 1.8\% \\
GPT4.1-mini & 17/427 & 4.0\% & 31/583 & 5.8\% & 23/342 & 6.7\% \\
Claude4-sonnet &5/389 &1.3\% & 17/453& 3.8\%  &  11/274& 4.0\% \\
Gemini2.5-pro & 0/336& 0.0\%&  18/461 &  3.9\% & 7/291 & 2.4\% \\
Gemini2.5-flash &3/466 & 0.6\%&  3/527& 0.6\% & 6/362 & 1.7\% \\
Deepseek-v3 & 4/356 &1.1\% &5/474  &1.1\% & 7/267 & 2.6\% \\
\bottomrule 
\end{tabular}
\label{hallucination}
\end{table}

The results in Table~\ref{hallucination} show that these models exhibit varying degrees of hallucination issues (i.e., generating spurious terms). For instance, the GPT4.1 model has the highest hallucination rate in the Trade Union Law (7.0\%), while other models also display hallucination issues to different extents. However, some models (such as Gemini2.5-pro) have lower hallucination rates (e.g., 0\% in the Standardization Law). This indicates that while these models demonstrate some effectiveness in multilingual terminology extraction, they still generate unsupported or erroneous terms in certain cases, highlighting the need for careful review and correction of the model outputs.
  
\section{Discussion}\label{sec12}
\begin{enumerate}

\item {\textbf{Why not adopt statistical machine translation (SMT) or co-occurrence-based translation tables for multilingual term extraction?}\
Although statistical machine translation (SMT) and co-occurrence-based translation tables have seen widespread use in bilingual terminology extraction, our experience—and a body of prior research—suggests that these methods are fundamentally unsuited to the complexities of legal language, especially when working with Chinese and Japanese. One persistent problem is that SMT relies on the existence of clear word boundaries and stable alignments, yet Chinese often lacks explicit segmentation, while Japanese features extensive compounding and a flexible, agglutinative structure. In practice, this means that statistical alignments are easily thrown off by ambiguous syntax or unseen word forms. Even with high-quality data, phrase or sentence-level alignments rarely yield the granularity required for reliable legal term mapping. Indeed, numerous studies have pointed out that for these language pairs, basic alignment itself remains a bottleneck—before we even get to the domain-specific nuances of law~\citep{kubota2002mostly,zhang-etal-2006-subword,koehn-etal-2003-statistical,koehn-knowles-2017-six}.}

\item {\textbf{Why is human-AI collaboration necessary—can't large language models do everything automatically?}\
Despite the remarkable progress of large language models (LLMs) in legal NLP, our work and direct testing reveal the limits of purely automated pipelines. While LLMs can process large volumes of text and generate plausible legal terminology suggestions, they are not immune to common pitfalls: errors in detecting term boundaries, contextual mismatches, and at times, outright hallucinations. These issues become especially pronounced in complex legal passages or in under-resourced language pairs, where training data is sparse and ambiguity is high. Our case studies repeatedly showed that—even with state-of-the-art models—redundancy and inconsistency can propagate through the extraction pipeline if left unchecked. This is why expert human review remains essential: not only to correct and clarify terminology, but to ensure that the results actually comply with legal and professional standards. In fields as sensitive as law, human-AI partnership is less a luxury than a necessity—crucial both for quality assurance and for meeting regulatory expectations.}

\item {\textbf{What is the academic contribution and originality of such an engineering-intensive workflow?}\
Rather than presenting yet another ``proof-of-concept,'' this study delivers a scalable, production-ready workflow for multilingual legal terminology mapping across Chinese, Japanese, and English law. To our knowledge, this is the first comprehensive system that integrates multi-agent AI automation with rigorous expert validation and a collaborative, open infrastructure. The technical demands involved here are not simply hurdles, but essential features that ensure the robustness and real-world utility of our approach. The originality lies not only in individual algorithms, but in the orchestration of human-machine synergy, the framework for ongoing data governance, and the infrastructure for continuous improvement. Building such a system required iterative problem solving, practical trade-offs, and sustained collaboration—underscoring both the complexity and necessity of this kind of work.}

\item {\textbf{How should criteria be defined, weighted, and validated to balance objectivity and context sensitivity?}\
Carefully constructed evaluation criteria are at the heart of reliable terminology extraction and mapping. In our framework, we introduced a five-dimensional scoring system—spanning Coverage, Consistency, Completeness, Professionalism, and Translation Quality—each broken down into 17 sub-criteria. We intentionally combined automated scoring with expert validation to strike a balance between objectivity and context sensitivity. That said, we acknowledge that our current weightings are an initial attempt, reflecting the priorities of Chinese, Japanese, and English legal domains as we see them. We expect that further empirical testing and input from other researchers will be needed to refine these rubrics for different domains or use cases. Our goal is to provide a transparent, adaptable baseline—open to critique, extension, and data-driven recalibration as the field evolves.}

\end{enumerate}

\section{Conclusion}\label{sec13}

This research puts forward a practical, human-AI collaborative framework designed to address the pressing need for scalable and reliable legal terminology resources—especially for less-resourced language pairs like Chinese and Japanese. Rather than relying on conventional manual approaches or purely automated tools, our workflow combines multi-agent automation with ongoing expert review, allowing for end-to-end extraction, alignment, and standardization of legal terms. In our empirical evaluation using a substantial trilingual legal corpus, this approach led to marked improvements in term coverage, semantic coherence, and contextual accuracy. The open, cloud-based ``Terminology-as-a-Service'' platform we developed further enables continuous quality management and collaborative curation. Interestingly, we also found that recent open-source large language models can perform at a level comparable to closed systems, suggesting that robust and cost-effective solutions for multilingual legal NLP are increasingly within reach. Our findings point to three main contributions. First, the hybrid human-AI methodology proved effective in tackling the structural, linguistic, and conceptual challenges that often undermine legal terminology mapping. Second, by combining quantitative measures with expert assessment, our evaluation framework provides a reproducible standard for future research on terminology quality. Third, by making the platform openly accessible and adaptable, we hope to foster ongoing collaboration and the sustainable growth of legal knowledge resources.

Nevertheless, this work is not without its challenges. Questions remain about how best to design and weight evaluation criteria, how to generalize the workflow to other legal systems and languages, and how to limit error propagation from automated modules. Addressing these issues will require further experimentation and input from the wider research community. We plan to make our multilingual legal terminology database and supporting platform available to the public soon, with the hope that it will spark broader collaboration and accelerate progress in this important field.

\backmatter

% \bmhead{Supplementary information}

% If your article has accompanying supplementary file/s please state so here. 

% Authors reporting data from electrophoretic gels and blots should supply the full unprocessed scans for key as part of their Supplementary information. This may be requested by the editorial team/s if it is missing.

% Please refer to Journal-level guidance for any specific requirements.

\bmhead{Funding}
This work is partly supported by the Humanities and Social Sciences Youth Pre-Research Project of East China Normal University (2022ECNU-YYJ062) and the National Natural Science Foundation of China Grant (No. 62306173).

% \section*{Declarations}

% Some journals require declarations to be submitted in a standardised format. Please check the Instructions for Authors of the journal to which you are submitting to see if you need to complete this section. If yes, your manuscript must contain the following sections under the heading `Declarations':

% \begin{itemize}
% \item Conflict of interest/Competing interests (check journal-specific guidelines for which heading to use)
% \item Ethics approval and consent to participate
% \item Consent for publication
% \item Materials availability
% \end{itemize}

\noindent
% If any of the sections are not relevant to your manuscript, please include the heading and write `Not applicable' for that section. 

% %%===================================================%%
% %% For presentation purpose, we have included        %%
% %% \bigskip command. Please ignore this.             %%
% %%===================================================%%
% \bigskip

% Springer journals and proceedings: \url{https://www.springer.com/gp/editorial-policies}

% Nature Portfolio journals: \url{https://www.nature.com/nature-research/editorial-policies}

% \textit{Scientific Reports}: \url{https://www.nature.com/srep/journal-policies/editorial-policies}

% BMC journals: \url{https://www.biomedcentral.com/getpublished/editorial-policies}
% \end{flushleft}

\begin{appendices}

\section{Detailed Statistics}\label{secA}
To evaluate the consistency of human annotators, we measured the ratings of five annotators on five evaluation dimensions (coverage, consistency, completeness, professionalism, and translation quality) using Cronbach's alpha and the two-way random effects Intraclass Correlation Coefficients ($ICC(2,1)$ and $ICC(2,k)$). The results show that, except for the ``professionalism'' dimension, the internal consistency across the dimensions is moderate: Coverage $\alpha=0.758$, Consistency $\alpha=0.727$, Completeness $\alpha=0.747$, and Translation Quality $\alpha=0.773$; the ``professionalism'' dimension has $\alpha=0.387$. Meanwhile, the $ICC(2,1)$ values are generally low (ranging from $0.015$ to $0.144$), indicating limited absolute consistency between individual annotators. However, when aggregating the mean ratings from multiple annotators, the consistency improves but remains low ($ICC(2,k)=0.070-0.458$). These results suggest that while annotators show some consistency in relative rankings, there are systematic differences in their rating scales, especially evident in the ``professionalism'' dimension.

\begin{table}[htbp]
\centering
\caption{Human annotator consistency statistics.}
\begin{tabular}{l|lll}
\toprule
\textbf{Metric}     & \textbf{Alpha} & \textbf{ICC(2,1)} & \textbf{ICC(2,k)} \\ \toprule
Coverage            & 0.758          & 0.061             & 0.244            \\  
Consistency         & 0.727          & 0.144             & 0.458            \\  
Completeness        & 0.747          & 0.062             & 0.248            \\  
Professionalism     & 0.387          & 0.015             & 0.070            \\  
Translation Quality & 0.773          & 0.072             & 0.280            \\ \bottomrule
\end{tabular}
\end{table}
\begin{table}[htbp]
\centering
\small
\caption{Correlations (Pearson $r$) and Correlations (Spearman $\rho$) between human 5-rater means and three LLMs (N=9). \footnotesize{Note: DeepSeek-v3 Coverage is constant (all 85; zero variance), so correlation is not defined (N/A).}}
\label{tab:llm-human-corr-pearson}
\begin{tabular}{l|rrrrr|r}
\toprule
\textbf{LLM} & Cov. & Cons. & Compl. & Prof. & Trans. & Overall \\
\midrule
DeepSeek-v3    &   N/A   & 0.646 & 0.739 & -0.019 & 0.439 & 0.820 \\
GPT4.1         & 0.386   & 0.860 & 0.585 & 0.409  & 0.170 & 0.842 \\
Gemini2.5-pro  & -0.663  & 0.828 & 0.696 & 0.790  & 0.611 & 0.875 \\
\bottomrule
\end{tabular} 
\begin{tabular}{l|rrrrr|r}
\toprule
\textbf{LLM} & Cov. & Cons. & Compl. & Prof. & Trans. & Overall \\
\midrule
DeepSeek-v3    &   N/A   & 0.687 & 0.769 & 0.073 & 0.414 & 0.667 \\
GPT4.1         & 0.419   & 0.838 & 0.725 & 0.471 & 0.177 & 0.828 \\
Gemini2.5-pro  & -0.498  & 0.833 & 0.640 & 0.848 & 0.731 & 0.800 \\
\bottomrule
\end{tabular}
\end{table}
Based on N=9 models, automatic scores from the three LLMs show the strongest and most stable correlations with the human 5-rater means on Overall. Strong to moderate positive correlations are also observed on Consistency and Completeness. Performance on Professionalism varies across models: Gemini2.5-pro aligns strongly with human ratings, GPT4.1 shows moderate alignment, while DeepSeek-v3 is weak. For Coverage, DeepSeek-v3's LLM scores are constant, making correlation undefined; Gemini2.5-pro exhibits a negative correlation, and GPT4.1 shows low-to-moderate positive correlation. Overall, LLM scores align well with human Overall and structural dimensions (Consistency/Completeness), whereas alignment on Coverage and Professionalism depends on the specific model and the rubric design.

\section{Statistics for the Chinese Law Corpus}\label{secB}
% A long table for vocabulary statistics for parallel corpus.
\begin{longtable}{>{\raggedright\arraybackslash}m{4cm}>{\centering\arraybackslash}m{1cm}>{\raggedleft\arraybackslash}m{1cm}>{\raggedleft\arraybackslash}m{1.5cm}>{\raggedleft\arraybackslash}m{1.5cm}>{\raggedleft\arraybackslash}m{1.5cm}}
\hline
\textbf{Law Name} & \textbf{Year} & \textbf{Entries} & \textbf{Chinese Words} & \textbf{Japanese Words} & \textbf{English Words} \\
\toprule
\endfirsthead
\toprule
\textbf{Law Name} & \textbf{Year} & \textbf{Entries} & \textbf{Chinese Words} & \textbf{Japanese Words} & \textbf{English Words} \\
\toprule
\endhead
Civil Code & 2021 & 1,400 & 54,717 & 89,603 & 78,891 \\
Labor Contract Law & 2007 & 112 & 5,325 & 8,272 & 8,146 \\ 
Criminal Procedure Law & 2012 & 334 & 16,343 & 28,891 & 25,139 \\
Copyright Law & 2020 & 84 & 5,328 & 8,620 & 7,359 \\
Tort Liability Law & 2010 & 107 & 3,798 & 5,947 & 5,280 \\
Foreign Investment Law & 2019 & 51 & 2,013 & 2,940 & 2,875 \\
Standardization Law & 2017 & 55 & 2,458 & 3,737 & 3,455 \\
Personal Info Protection Law & 2021 & 88 & 4,498 & 6,985 & 5,998 \\
Against Unfair Competition Law & 2019 & 41 & 2,280 & 3,488 & 3,194 \\
Advertising Law & 2021 & 83 & 5,057 & 8,161 & 6,374 \\
Patent Law & 2020 & 93 & 5,551 & 8,890 & 8,538 \\
Labor Dispute Mediation and Arbitration Law & 2008 & 65 & 2,999 & 4,159 & 4,118 \\
Administrative Penalty Law & 2021 & 101 & 4,838 & 8,268 & 7,668 \\
Exit and Entry Administration Law & 2012 & 108 & 6,230 & 9,427 & 8,337 \\
Anti-Monopoly Law & 2022 & 81 & 4,036 & 6,099 & 5,570 \\
Science and Technology Progress Law & 2007 & 86 & 4,223 & 6,490 & 6,316 \\
Statistics Law & 2009 & 60 & 2,874 & 4,859 & 4,564 \\
Coast Guard Law & 2021 & 98 & 4,953 & 7,403 & 6,653 \\
Labor Law & 1995 & 123 & 3,993 & 6,364 & 5,909 \\
Intangible Cultural Heritage Law & 2011 & 54 & 2,441 & 3,389 & 3,349 \\
Criminal Law & 2011 & 509 & 29,183 & 51,785 & 47,362 \\
Seed Law & 2021 & 105 & 7,023 & 11,267 & 11,543 \\
Transformation Promotion Law of Scientific and Technological Achievements & 2015 & 61 & 3,235 & 5,015 & 4,853 \\
Anti-Espionage Law & 2023 & 80 & 4,053 & 6,719 & 6,117 \\
Circular Economy Promotion Law & 2008 & 68 & 3,984 & 5,996 & 5,408 \\
Social Insurance Law & 2011 & 113 & 4,962 & 7,365 & 7,298 \\
Constitution & 2018 & 162 & 7,504 & 11,770 & 11,823 \\
Data Security Law & 2021 & 66 & 2,726 & 4,196 & 3,961 \\
Renewable Energy Law & 2009 & 44 & 2,355 & 3,848 & 3,358 \\
Company Law & 2023 & 293 & 15,461 & 24,566 & 22,077 \\
Trademark Law & 2020 & 85 & 5,607 & 9,169 & 8,130 \\
E-Commerce Law & 2018 & 102 & 4,976 & 7,202 & 6,601 \\
Trade Union Law & 2001 & 67 & 2,886 & 5,019 & 4,744 \\
Work Safety Law & 2021 & 129 & 9,495 & 15,190 & 14,474 \\
Foreign-Related Civil Relations Application Law & 2010 & 64 & 2,000 & 2,426 & 2,381 \\
\bottomrule
\textbf{Total} & & \textbf{5,172} & \textbf{249,405} & \textbf{403,525} & \textbf{367,863} \\
\bottomrule
\caption{Vocabulary Statistics for Chinese laws}
\end{longtable}

\section{Samples of the Extracted Term Mappings}\label{secC}
\begin{landscape}
\scriptsize
\begin{longtable}{p{1cm}|p{1cm}|p{1cm}|p{2cm}|p{4cm}|p{3cm}|p{2cm}|p{0.8cm}|p{1.2cm}}
\toprule
\textbf{Chinese} & \textbf{Japanese} & \textbf{English} & \textbf{Chinese Context} & \textbf{English Context} & \textbf{Japanese Context} & \textbf{Explanation}  & \textbf{Art.ID} & \textbf{Law} \\
\toprule
被侵权人	&権利被侵害者	&victim of a tort &	被侵权人有权请求侵权人承担侵权责任&	The victim of a tort shall be entitled to require the tortfeasor to assume the tort liability &権利被侵害者は、権利侵害者に対し権利侵害責任を負うよう請求する権利を有する。&   	指因他人侵权行为而受到损害的人; 权利受到侵害的一方，当事人 & 3 & 中华人民共和国侵权责任法(2010) \\	
\midrule
间谍行为	&スパイ行為	&espionage&	涉及间谍行为的网络信息内容&	certain network information involving espionage&	スパイ行為に関わるネットワーク情報コンテンツ	&秘密获取国家机密或情报的行为; 秘密收集、传递国家机密或情报的违法行为&	 36&		中华人民共和国反间谍法(2023)\\	
\midrule
摄制权	&撮影製作権	&right of cinematography&摄制权，即以摄制视听作品的方法将作品固定在载体上的权利& the right of cinematography, that is, the right to fix a work on the medium by producing an audiovisual work	&撮影製作権、すなわち視聴覚著作物の撮影製作方法により、著作物を媒体上に固定させる権利。		&以摄制视听作品方式固定作品的权利; 通过摄制方式将作品固定在载体上的权利	& 10&	中华人民共和国著作权法(2020)\\	
\midrule
金属冶炼建设项目&	金属精錬建設プロジェクト	&construction projects for metal smelting	&矿山、金属冶炼建设项目和用于生产、储存、装卸危险物品的建设项目的安全设施设计应当按照国家有关规定报经有关部门审查&	construction projects for mines and metal smelting and construction projects for the manufacture, storage, loading and unloading of dangerous articles shall, in accordance with relevant state regulations, be submitted to relevant departments for examination&	鉱山および金属精錬建設プロジェクトおよび危険物の生産、貯蔵または積み下ろしに用いる建設プロジェクトの安全施設の設計については、国の関連規定に従い関連機関に報告し審査を経なければならない。	&	涉及金属冶炼工艺的建设工程项目; 涉及金属冶炼工艺的建设项目& 33	& 	中华人民共和国安全生产法(2021)\\	
\midrule
构成侵权的初步证据&	権利侵害となることの一次的な証拠&	preliminary evidence establishing the tort&	通知应当包括构成侵权的初步证据及权利人的真实身份信息	&The notice shall include the preliminary evidence establishing the tort and the real identity information of the right holder&	権利侵害となることの一次的な証拠及び権利者の真実の身分情報を含むものでなければならない	&证明侵权行为存在的初步证据材料& 1195	 &  中华人民共和国民法典(2021)\\	
\midrule
劳动者	&勤労者	&working people&	中华人民共和国劳动者有休息的权利	&Working people in the People's Republic of China shall have the right to rest&	中華人民共和国の勤労者は休息の権利を有する	&指在中华人民共和国境内从事劳动活动的人员; 指在劳动关系中提供劳动的个人，包括职工和其他劳动人员	&	43 &	中华人民共和国宪法(2018)\\	
\midrule
专利申请文件	&専利出願書類	&patent application documents&	收到专利申请文件之日为申请日	&the patent application documents are received	&専利出願書類を受領した日		&提交申请专利所需的书面材料; 提交申请专利保护所需的正式文件& 28&中华人民共和国专利法(2020)\\	
\midrule	
上市公司	&上場会社	&listed companies	&上市公司应当依法披露股东、实际控制人的信息	&Listed companies shall disclose information on shareholders and actual controllers in accordance with the law&	上場会社は、法により株主、実質的支配者の情報を開示しなければならず		&在证券交易所公开发行股票的公司; 在证券交易所公开发行股票并上市交易的公司		& 140 &	中华人民共和国公司法(2023)\\	
\midrule	
遗产份额	&遺産相続分	&portion of an estate&	保留必要的遗产份额&	Reservation of a necessary portion of an estate	&必要な遺産相続分を留保すること	& 指继承人依法应得的遗产部分; 继承人依法应得的遗产比例或部分	&	 1141&	中华人民共和国民法典(2021)\\	
\midrule
上缴国库		&国庫に納入する	&	turned over to the state treasury		&一律上缴国库	&	shall be turned over to the state treasury		&国庫に納入する	&		将财物交付国家财政部门管理; 将财物交付国家财政部门管理		& 234	&中华人民共和国刑事诉讼法(2012)\\	
\midrule
出境入境证件&	出入国証書&exit/entry documents	&未持有效出境入境证件	&Hold no valid exit/entry documents	&有効な出入国証書を持たない	& 用于证明公民合法出入境身份的官方证件	& 12	&中华人民共和国出境入境管理法(2012)\\
\bottomrule
\end{longtable}
\end{landscape}

\end{appendices}


\bibliography{sn-bibliography}% common bib file

\end{document}